\pdfoutput=1

\documentclass[11pt]{article}

\usepackage[final]{acl}

\usepackage{times}
\usepackage{latexsym}

\usepackage{textcomp}
\usepackage{newtxtext}

\usepackage[T1]{fontenc}

\usepackage[utf8]{inputenc}

\usepackage{microtype}

\usepackage{inconsolata}

\usepackage{graphicx}

\usepackage{inconsolata}
\usepackage{amssymb}
\newenvironment{dotpar}{\noindent\textbullet\ }{\par\vspace{\parskip}}
\usepackage{booktabs}
\usepackage{tabularx}
\usepackage{multirow}
\usepackage{xcolor}
\usepackage{hyperref}
\usepackage{amssymb}
\usepackage{pifont}

\newcommand{\xmark}{\ding{55}}
\newcommand{\cmark}{\ding{51}}

\title{Benchmarking Chinese Commonsense Reasoning of LLMs: From Chinese-Specifics to Reasoning-Memorization Correlations}

\author{
 \textbf{Jiaxing Sun\textsuperscript{1}\thanks{Equal contribution.}},
 \textbf{Weiquan Huang\textsuperscript{2}\footnotemark[1]},
 \textbf{Jiang Wu\textsuperscript{3}\footnotemark[1]\thanks{Project lead.}},
 \textbf{Chenya Gu\textsuperscript{3}},
 \textbf{Wei Li\textsuperscript{3}},
\\
 \textbf{Songyang Zhang\textsuperscript{3}},
 \textbf{Hang Yan\textsuperscript{3}},
 \textbf{Conghui He\textsuperscript{3}\thanks{Corresponding author.}},
\\
\\
 \textsuperscript{1}State Key Laboratory of Information Engineering in Surveying, Mapping and Remote Sensing, \\Wuhan University \\
 \textsuperscript{2}Tongji University  
 \textsuperscript{3}Shanghai AI Laboratory,
\\
 \small{
   \textbf{Correspondence:} \href{heconghui@pjlab.org.cn}{heconghui@pjlab.org.cn}
 }
}

\begin{document}
\maketitle
\begin{abstract}

We introduce CHARM, the first benchmark for comprehensively and in-depth evaluating the commonsense reasoning ability of large language models (LLMs) in Chinese, which covers both globally known and Chinese-specific commonsense. We evaluated 7 English and 12 Chinese-oriented LLMs on CHARM, employing 5 representative prompt strategies for improving LLMs' reasoning ability, such as Chain-of-Thought. Our findings indicated that the LLM's language orientation and the task's domain influence the effectiveness of the prompt strategy, which enriches previous research findings.
We built closely-interconnected reasoning and memorization tasks, and found that some LLMs struggle with memorizing Chinese commonsense, affecting their reasoning ability, while others show differences in reasoning despite similar memorization performance. We also evaluated the LLMs' memorization-independent reasoning abilities and analyzed the typical errors. Our study precisely identified the LLMs' strengths and weaknesses, providing the clear direction for optimization. It can also serve as a reference for studies in other fields. We will release CHARM at \url{https://github.com/opendatalab/CHARM}.
\end{abstract}

\begin{figure*}[!t]
    \centering
    \includegraphics[width=1\linewidth]{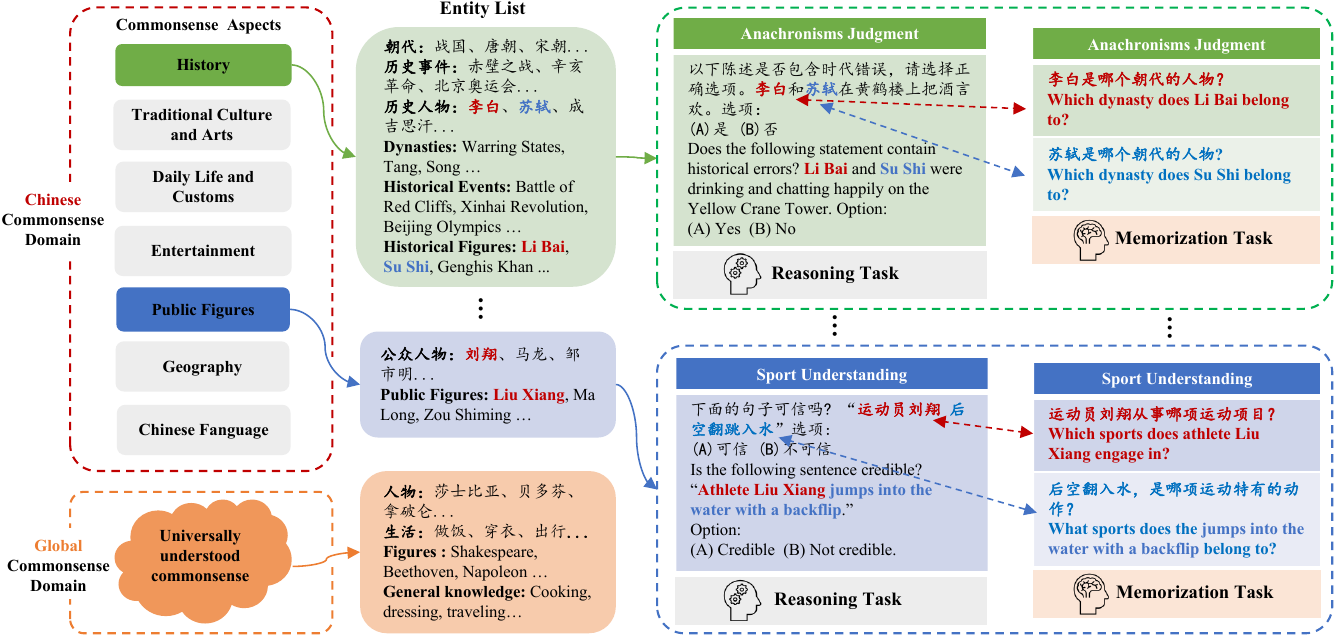}
    \caption{Construction of CHARM. CHARM encompasses both global and Chinese-specific commonsense. CHARM consists closely-interconnected reasoning and memorization tasks.}
    \label{fig:fig2_CHARM_construction}
\end{figure*}

\begin{table*}[h]
\small
\centering
\begin{tabular}{p{5cm}p{1.5cm}p{1.0cm}p{2.0cm}p{2.0cm}p{1.5cm}}
\toprule
\textbf{Benchmarks} & \textbf{CN-Lang} & \textbf{CSR} & \textbf{CN-specifics} & \textbf{Dual-Domain} & \textbf{Rea-Mem} \\
\midrule
{Most benchmarks in \cite{davis2023benchmarks}}  & \xmark & \cmark & \xmark &\xmark & \xmark \\
XNLI, XCOPA, XStoryCloze & \cmark & \cmark & \xmark & \xmark &\xmark \\
LogiQA, CLUE, CMMLU & \cmark & \xmark & \cmark & \xmark  & \xmark  \\
CORECODE & \cmark & \cmark &\xmark & \xmark & \xmark  \\
\midrule
\textbf{CHARM (ours)} & \cmark & \cmark & \cmark &\cmark &\cmark\\
\bottomrule
\end{tabular}
\caption{Comparison of commonsense reasoning benchmarks. ``CN-Lang'' indicates the benchmark is presented in Chinese language. ``CSR'' means the benchmark is designed to focus on \textbf{C}ommon\textbf{S}ense \textbf{R}easoning. ``CN-specific'' indicates the benchmark includes elements that are unique to Chinese culture, language, regional characteristics, history, etc. ``Dual-Domain'' indicates the benchmark encompasses both Chinese-specific
and global domain tasks, with questions presented in the similar style and format. ``Rea-Mem'' indicates the benchmark includes closely-interconnected \textbf{rea}soning and \textbf{mem}orization tasks.}
\label{tab:benchmark_comparison}
\end{table*}

\section{Introduction}

Commonsense reasoning is important for the enhancement of the large language models (LLMs) \cite{bommasani2021foundation,GPT4} towards artificial general intelligence (AGI) \cite{davis2015commonsense}, therefore requires thorough evaluations.
Numerous benchmarks evaluate the commonsense reasoning of LLMs, but most are English-based, limiting non-English evaluations \cite{davis2023benchmarks}. This paper focuses on assessing LLMs' commonsense reasoning in a Chinese context. Currently, some commonsense reasoning benchmarks in Chinese are simply English translations \cite{conneau2018xnli,ponti2020xcopa,lin2022few}, which overlooks unique Chinese cultural, linguistic, regional, and historical aspects. These factors matter when Chinese users use the LLM, hence should be included in benchmarks. To effectively tackle this, we introduce CHARM, the benchmark designed to thoroughly and in-depth assess the abilities of LLMs in Chinese commonsense reasoning. It covers two domains: globally accepted commonsense (global domain) and Chinese-specific commonsense (Chinese domain). The latter includes 7 aspects: \textit{History (H)}, \textit{Traditional Culture and Arts (CA)}, \textit{Daily Life and Customs (LC)}, \textit{Entertainment (E)}, \textit{Public Figures (F)}, \textit{Geography (G)}, and \textit{Chinese Language (L)}. Therefore CHARM allows a thorough evaluation of LLMs' reasoning in a Chinese context.

Prompt strategies like Chain of Thought (CoT) \cite{wei2022cot} can significantly improve LLMs' reasoning performance \cite{wang2022self-con,wang2023plan-and}. Particularly, as the training corpus of LLMs is primarily in English \cite{touvron2023llama}, studies \cite{shi2022EN-cot,huang2023XLT,zhang2023donTrustChatGPT} have shown that for non-English reasoning tasks, some LLMs perform better when reasoning in English than the native language.
We evaluated 7 English and 12 Chinese-oriented LLMs on CHARM, employing 5 representative prompt strategies. The result showed that prompt strategies' effectiveness depends on the LLMs' orientation and the benchmark task's domain, which enriches prior research and guides performance assessment and strategy choice for non-English LLMs.

LLMs' commonsense reasoning relies on memorization. Exploring the correlation between memorization and reasoning offers insights into LLMs, aiding deeper understanding and suggesting ways to enhance these abilities\cite{bian2023chatgptKnowledgeable}. Some benchmarks \cite{yu2023kola,wang2023seaeval,fei2023lawbench} aid the research of memorization-reasoning relationships by incorporate tasks for assessing knowledge memorization and application (like reasoning). However, they used the existing and disparate datasets for different tasks, resulting in a lack of intrinsic connections between these tasks. 
For instance, the question $Q_{rea}$ tests the LLM's reasoning with the knowledge piece $K$. However, in memorization tasks, there probably is not any matching questions to determine if the LLM has effectively memorized $K$. 
Hence, if the LLM fails on $Q_{rea}$, it's unclear whether due to poor reasoning or forgetfulness of $K$. This results in the disjointed evaluation of memorization and reasoning, failing to uncover their intrinsic links. To address this limitation, we selected suitable reasoning tasks from CHARM's Chinese domain, and built related memorization questions for each reasoning question (see Figure \ref{fig:fig2_CHARM_construction}). 
This design produces the closely-interconnected reasoning and memorization tasks, therefore allows for not only the concurrent evaluation of the two abilities, but also the assessment of memorization-independent reasoning, providing the clear guidance for the LLMs' enhancement.

The contributions of this paper are as follows:
\begin{itemize}
    \item We present CHARM, the first benchmark for comprehensively evaluating the LLMs' commonsense reasoning ability in Chinese, by encompassing not only the global but also the Chinese-specific commonsense.
\end{itemize}

\begin{itemize}
    \item We evaluated the representative prompt strategies on CHARM. Results showed that LLMs' orientation and the task's domain affect prompt strategy performance, which enriches previous research findings.
\end{itemize}

\begin{itemize}
    \item In CHARM, we built closely-interconnected reasoning and memorization tasks in Chinese commonsense domain, allowing for in-depth understanding the correlation between these abilities and precisely identifying the LLMs' strengths and weaknesses. The design approach could serve as the reference for other fields.
\end{itemize}

\section{Related Work}

\textbf{Commonsense Reasoning Benchmarks  }
There are lots of commonsense reasoning benchmarks, most of them are in English \cite{davis2023benchmarks}. 
Some Chinese commonsense reasoning benchmarks are directly translated from English benchmarks \cite{conneau2018xnli,ponti2020xcopa,lin2022few}, which lack the Chinese specifics. 
There are some native Chinese benchmarks that include some Chinese-specific factors and involve commonsense reasoning to a certain extent, such as LogiQA \cite{liu2020logiqa1.0,liu2023logiqa2.0}, CLUE \cite{xu2020clue} and CMMLU \cite{li2023cmmlu}. However, they are not designed for commonsense reasoning, therefore containing the large portion of irrelevant tasks and questions.
CORECODE \cite{shi2023corecode} is the benchmark for Chinese commonsense reasoning and commonsense conflict detection, but it is not strictly designed to distinguish the Chinese-specific and global domains when compared with CHARM. In addition, CHARM has the closely-interconnected reasoning and memorization tasks, which are not included in previous commonsense reasoning benchmarks.
The comparison of CHARM with previous commonsense reasoning benchmarks is shown in Table \ref{tab:benchmark_comparison}.

\textbf{Prompt Strategy  }
Prompt strategies such as CoT \cite{wei2022cot} can effectively boost the reasoning capabilities of LLMs \cite{wang2022self-con,wang2023plan-and}. Notably, as the LLM training corpus is primarily in English \cite{touvron2023llama}, research revealed that for reasoning tasks in non-English languages, some LLMs exhibit superior performance when reasoning in English as opposed to the native language \cite{shi2022EN-cot,zhang2023donTrustChatGPT,huang2023XLT}.
\cite{kim2023IN-CLT} proposed a novel cross-language transfer prompt method, which uses both the source and target languages to construct examples.

\textbf{Benchmarks on Correlations of Memorization and Reasoning  }
There are benchmarks which assess both the knowledge memorization and reasoning capabilities of the LLMs within specific domains. 
For instance, KoLA \cite{yu2023kola}, with its focus on world knowledge, includes tasks related to knowledge memorization and application (reasoning).
SeaEval \cite{wang2023seaeval}, emphasizing cross-language consistency and multicultural reasoning, involves tasks for cultural understanding and complex reasoning. 
There are also benchmarks aimed at specialized fields, like LawBench \cite{fei2023lawbench}, which include tasks for both memorization and application.

\section{CHARM}

CHARM is built for comprehensive and in-depth evaluation of LLMs in Chinese commonsense reasoning and revealing the intrinsic correlation between memorization and reasoning. Therefore, CHARM covers two domains, global and Chinese, using carefully selected tasks for comprehensive coverage. In addition, we chose reasoning tasks and constructed the closely-tied memorization tasks.
The construction and main features of CHARM are in Figure \ref{fig:fig2_CHARM_construction}. The detailed composition of CHARM is in Table \ref{tab:charm_overview}.

\subsection{Commonsense Domain}
\label{sec:commonsense-domain}

\textbf{Global commonsense domain} consists of universally understood commonsense.
It covers objects and aspects of modern life that an individual should be aware of.
It includes foundational knowledge that someone with a basic modern education is expected to know.  
When it involves individuals, they are globally recognized figures.

\textbf{Chinese commonsense domain} encompasses Chinese-specific elements. We categoried them into 7 aspects:

\textbf{\textit{History (H)}} includes important events and figures in Chinese history, China's dynasties, and other basic  facts and shared knowledge about the history of China.

\textbf{\textit{Traditional Culture and Arts (CA)}} encompasses Chinese traditional cultural arts, literary works, and traditional lifestyles.

\textbf{\textit{Daily Life and Customs (LC)}} includes modern Chinese daily routines, clothing, food, housing, transportation festivals and so on.

\textbf{\textit{Entertainment (E)}} includes the movies, television programs, music, and other entertainments in modern Chinese daily life.

\textbf{\textit{Public Figures (F)}} encompasses the public figures well-known in Chinese society.

\textbf{\textit{Geography (G)}} includes China's geographical distribution, natural landscapes, and characteristic regional cultures.

\textbf{\textit{Chinese Language (L)}} includes the fundamentals of the Chinese language, such as Chinese characters, idioms and so on.

For the two domains, especially for the above 7 aspects, we collected corresponding entities, forming the lists as shown in the Figure \ref{fig:fig2_CHARM_construction} and \ref{fig:entity-and-question-examples-Chinese-commonsense-aspects}. 
Most of the entities were selected from Gaokao Bench\footnote{Gaokao Bench is the collection of China's university entrance exam questions, which contributes to all the 7 aspects.}\cite{zhang2023gaokaobench}, Douban\footnote{\url{https://www.douban.com/} is the popular user-centric cultural review platform in China, which mainly contributes to the \textit{Entertainment} aspect.}, Hupu\footnote{\url{https://www.hupu.com/} is the large sports community popular in China, which mainly contributes to the \textit{Public Figures} aspect.}. Some entities were collected with the help of searching engines. We only collected the entities that are well-known in China.
These entities were then used to create the commonsense reasoning questions, which belong to the corresponding domain and aspect.

\begin{table*}[!t]
    \centering
\tiny
\setlength\tabcolsep{6pt}
\begin{tabular*}{0.88\textwidth}
{p{1.2cm}|c|c|c|c|c|cc}
\toprule
 \textbf{Task Type} & \textbf{Task} & \textbf{Domain}  & \textbf{Chinese Aspects} & \textbf{Construction} & \textbf{Question Type} & \textbf{\# Question} \\
\midrule
\multirow{14}{*}{\textbf{Reasoning}} &
\multirow{2}{*}{Anachronisms Judgment (AJ)} & Chinese  &  \textit{H, AC, LC, F} &  [H] &  2-option MCQ &  150 \\ &
     &  global  &                    - & [T][H] &    2-option MCQ &   150 \\ &
       \multirow{2}{*}{ Time Understanding (TU)} & Chinese  &  \textit{H, AC, LC} &    [H] &    4-option MCQ &   100 \\ &
         &  global &                    - &    [T] & 5or6-option MCQ &   100 \\ &
    \multirow{2}{*}{Sequence Understanding (SqU)} & Chinese  &      \textit{H, CA, LC, G, L} &    [H] &    4-option MCQ &   100 \\ &
    &  global  &                    - & [T][H] &    4-option MCQ &   100 \\ &
\multirow{2}{*}{Movie and Music Recommendation (MMR)} & Chinese  &                 \textit{E} &    [H] &    4-option MCQ &    50 \\ &
 &  global  &                    - &    [T] &    4-option MCQ &    50 \\ &
          \multirow{2}{*}{ Sport Understanding (SpU)} & Chinese  &                    \textit{F} &    [H] &    2-option MCQ &   200 \\ &
           &  global  &                    - &    [H] &    2-option MCQ &   200 \\ &
    \multirow{2}{*}{Natural Language Inference (NLI)} & Chinese  &          \textit{G, E, L}, & [S][H] &    3-option MCQ &   100 \\ &
     &  global  &                    - &    [S] &    3-option MCQ &   100 \\ &
         \multirow{2}{*}{Reading Comprehension (RC)} & Chinese  &   all 7 aspects &    [S] &    4-option MCQ &   200 \\ &
         &  global  &   - &    [S] &    4-option MCQ &   200 \\ 
\midrule
\multirow{4}{*}{\textbf{Memorization}} & 
         Anachronisms Judgment (AJ) & Chinese  &        \textit{H,AC,LC,F} &    [H] &     Free-form QA &   150 \\ & 
            Time Understanding (TU) & Chinese  &        \textit{H, AC, LC} &    [H] &     Free-form QA &    83 \\ & 
Movie and Music Recommendation (MMR) & Chinese &                \textit{E} &    [H] &     Free-form QA &   399 \\ & 
           Sport Understanding (SpU) & Chinese &                \textit{F} &    [H] &     Free-form QA &   127 \\ 
\bottomrule
\end{tabular*}
\normalsize
    \caption{Overview of CHARM. The question numbers of reasoning and memorization tasks are 1800 and 759.}
    \label{tab:charm_overview}
\end{table*}

\subsection{Reasoning Tasks}
\label{sec:reasoning_task_setup}

When designing the reasoning tasks in CHARM, we beared two criteria in mind. First, the tasks should span both commonsense domains, particularly the 7 Chinese aspects. Second, the global and Chinese tasks should have identical types and settings, differing only in their commonsense domains. From the existing English commonsense reasoning datasets \cite{davis2023benchmarks,suzgun2022big-bench}, we selected the following 7 tasks:

\textbf{Anachronisms Judgment (AJ)} necessitates the LLM to identify anachronisms in provided sentences. This involves the LLM understanding the era associated with well-known figures, items, and events to facilitate commonsense-based reasoning. Global domain questions are the mix of translations\footnote{\url{https://github.com/google/BIG-bench/tree/main/bigbench/benchmark_tasks/anachronisms}} and handcrafted, while all Chinese domain questions are handcrafted.

\textbf{Time Understanding (TU)} requires the LLM infers a time (including year, date, moment, etc.) based on a given context, which necessitates the fundamental understanding of time-related commonsense and the capacity for mathematical reasoning. All question in the global domain are translations\footnote{\url{https://github.com/google/BIG-bench/tree/main/bigbench/benchmark_tasks/date_understanding}} and all in Chinese domain are handcrafted.

\textbf{Sequence Understanding (SqU)} requires the LLM sort a series of entities according to time or occurrence order, requiring logical reasoning based on commonsense. The global domain questions are the mix of translations\footnote{\url{https://github.com/google/BIG-bench/tree/main/bigbench/benchmark_tasks/logical_sequence}} and handcrafted; while all in the Chinese domain are handcrafted.

\textbf{Movie and Music Recommendation (MMR)} necessitates the LLM identifies the most similar matches to a variety of movies or music tracks, requiring the understanding of these popular movies and music and ability to identify their commonalities. All global domain questions are translations\footnote{\url{https://github.com/google/BIG-bench/tree/main/bigbench/benchmark_tasks/movie_recommendation}}, and all in the Chinese domain are handcrafted.

\textbf{Sport Understanding (SpU) } involves a crafted sentence with a famous athlete and a common sport action, and the LLM must assess its credibility, which demands understanding of sports and commonsense judgement. The questions in both domains are handcrafted, refering \cite{suzgun2022big-bench} .

\textbf{Natural Language Inference (NLI)} gives two sentences and asks the LLM to classify their relationship as entailment, contradiction, or neutral, necessitating commonsense-based reasoning and judgement. All global domain questions are selected from CLUE \cite{xu2020clue}; the questions in the Chinese domain are partly from CLUE, and partly handcrafted.

\textbf{Reading Comprehension (RC)} gives a passage of text, and the LLM is required to reason based on it. All question in both domains are selected from LogiQA \cite{liu2020logiqa1.0,liu2023logiqa2.0}.

The chosen tasks adequately cover both the commonsense domains, particularly the 7 aspects of the Chinese commonsense domain. This coverage enables a comprehensive assessment of LLMs' commonsense reasoning ability in Chinese. Moreover, the Chinese-domain questions could be created following the similar types and settings as their global counterpart, facilitating the cleaner comparison of the LLMs’ performance across the domains.

All questions in the CHARM reasoning tasks are multiple-choice questions. Detailed information is in Table \ref{tab:charm_overview}.
Question examples of the tasks are in Figure \ref{fig:task_rea_examples} in Appendix \ref{sec:app_examples_rea_tasks}. We used regular expressions to extract the preferred choice from the generation of the LLMs \cite{huang2023c-eval,li2023cmmlu} and used \textit{accuracy} as the metric.

\subsection{Construction of Reasoning Tasks}

The construction of CHARM reasoning tasks involved the following three methods: 

\textbf{Translation [T]} was applied to some global domain reasoning tasks. We translated the English commonsense reasoning benchmarks mentioned in \textsection\ref{sec:reasoning_task_setup} using GPT-3.5. Then we replaced the English names with commonly used Chinese names and manually screen the translated questions, retaining those without translation errors and accepted as commonsense globally.

\textbf{Selection [S] } We selected the excellent native Chinese datasets, LogiQA \cite{liu2020logiqa1.0}, and CLUE \cite{xu2020clue}, and chose the questions that meet the requirements for CHARM.

\textbf{Handcraft [H] } was mainly applied to the most Chinese domain reasoning tasks. We used the entities in \textsection \ref{sec:commonsense-domain}, and referred to the corresponding global domain task questions (from [T] or [S]) to construct questions with the same type and style. This ensured that the same reasoning task in two domains only differs in the commonsense domain, thus facilitating cleaner comparative analysis, as shown in Figure \ref{fig:task_rea_examples}.

Detailed construction information of all reasoning tasks are shown in Table \ref{tab:charm_overview}.

\subsection{Memorization Tasks}

Shared commonsense knowledge pieces serve as links between reasoning and memorization questions. From the 7 reasoning tasks, we chose 4 that can be readily associated in this manner, \textbf{AJ}, \textbf{TU}, \textbf{MMR}, \textbf{SpU}, referred as the \textit{Memorization-Reasoning-Interconnected (MRI) tasks}, and built the related memorization questions.

\textbf{Construction }
We first extracted the commonsense knowledge pieces related to the entities in the corresponding reasoning questions. Information about each entity was collected to the degree sufficient to address the associated reasoning question, and then used to formulate the memorization questions. Following the \textit{Knowledge Memorization} task in KoLA \cite{yu2023kola}, we chose free-form QA instead of multiple-choice or true/false questions, which can effectively avoid the impact of randomness. All memorization questions were \textbf{handcrafted [H]}.
Question examples are in Figure \ref{fig:task_mem_example} in Appendix \ref{sec:app_examples_mem_tasks}. 
The averaged number of related memorization questions for each reasoning question are shown in Table \ref{tab:num_related_mem_per_rea}.

\begin{table}[t!]
\tiny
    \centering
    \begin{tabular}{ccccc}
    \toprule
       \textbf{Task} &\textbf{AJ} & \textbf{TU} & \textbf{SpU} & \textbf{MMR}  \\
    \midrule
        Avg. \# related memorization questions &2.1 & 3.2 & 2.0 & 8.0  \\
        \bottomrule
    \end{tabular}
    \caption{Averaged number of related memorization questions per reasoning question for each task.}
    \label{tab:num_related_mem_per_rea}
\end{table}

\textbf{Judgement and Metric } For the memorization task of \textbf{MMR}, we used a rule-based matching method for evaluation; for the other three tasks, we used GPT-3.5 for judgement. We used \textit{accuracy} as the metric.

\begin{figure}
    \centering
    \includegraphics[width=0.5\linewidth]{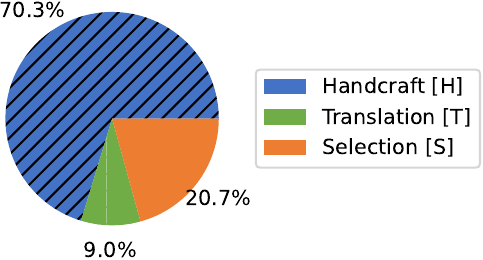}
    \caption{Distribution of CHARM construction.}
    \label{fig:construction_pie}
\end{figure}

\subsection{Quality Assurance}

The distribution of the contruction methods for CHARM is shown in Figure \ref{fig:construction_pie}. 
After construction of CHARM, we conducted the quality assurance to ensure the quality of the questions. We hired professional NLP annotators to review the questions. The quality assurance process involved five steps: (1) We prepared and assigned annotation task packages to annotators; (2) We trained annotators, emphasizing the avoidance of social bias; (3) We conducted a trial review on a random 20\% of questions to fine adjust the review process; (4) Two annotators independently reviewed each question and provided answers without seeing our answers, and a question passed only if both annotators' answers matched ours and they found no issues with the question; (5) For questions that failed in step 4, authors discussed whether to retain, discard, or correct them based on the nature of the issues identified.

Details about the quality assurance are in Appendix \ref{sec:annotation_details}.

\section{Experimental Setup}
\subsection{Language Models}

We evaluated the currently commonly used LLMs, which can be divided into two categories: (1) 7 English LLMs, including GPT series \cite{GPT4}, LLaMA-2 \cite{touvron2023llama2}, and Vicuna\footnote{\url{https://huggingface.co/lmsys/vicuna-7b-v1.5-16k} and \url{https://huggingface.co/lmsys/vicuna-13b-v1.5-16k}}. (2) 12 Chinese-oriented LLMs, including ChatGLM3\footnote{\url{https://huggingface.co/THUDM/chatglm3-6b-32k}}, Baichuan2 \cite{yang2023baichuan2}, InternLM2 \cite{team2023internlm}, Yi\footnote{\url{https://github.com/01-ai/Yi}}, DeepSeek\cite{bi2024deepseek} and Qwen \cite{bai2023qwen}. For open-source models, we chose the chat version instead of the base version. For closed-source models, we used the official API\footnote{We used the gpt-3.5-turbo-1106 version for GPT-3.5 and the gpt-4-1106-preview version for GPT-4.}. Detailed information is in Table \ref{tab:language_models}.

We used opencompass\footnote{\url{https://github.com/open-compass/opencompass}} in all our experiments. For all LLMs, the maximum out length was set to 512. For all open-source LLMs, we used the default settings in opencompass: the decoding temperature was the default value of the huggingface transformers library\footnote{\url{https://github.com/huggingface/transformers}}, which is 1.0; \textit{do\_sample} was set to False; the PyTorch numerical type was bf16. For closed-source models (GPT-3.5 and GPT-4), we used the default settings in opencompass: the temperature was set to 0.7.

\begin{table}
    \centering
    \tiny
    \begin{tabular*}{0.45\textwidth}{@{\extracolsep{\fill}}lccc}
    \toprule
\textbf{Models} & \textbf{Open Source?} & \textbf{Model Size} & \textbf{Primary Language} \\
    \midrule
LLaMA-2 &  Yes & 7B, 13B, 70B & English \\
Vicuna &  Yes & 7B, 13B & English \\
GPT-3.5 &  No & undisclosed & English \\
GPT-4 &  No & undisclosed & English \\
     \midrule
ChatGLM3 &  Yes & 6B & Chinese \\
Baichuan2 &  Yes & 7B, 13B & Chinese \\
InternLM2 &  Yes & 7B, 20B & Chinese \\
Yi &  Yes & 6B, 34B & Chinese \\
DeepSeek &  Yes & 7B, 67B & Chinese \\
Qwen &  Yes & 7B, 14B, 72B & Chinese \\    
    \bottomrule
    \end{tabular*}
    \normalsize
    \caption{LLMs evaluated in our experiments}
    \label{tab:language_models}
\end{table}

\begin{table*}
    \centering
\tiny
\begin{tabular*}{1.0\textwidth}{@{\extracolsep{\fill}}p{1.9cm}p{0.3cm}p{0.3cm}p{0.3cm}p{0.3cm}p{0.3cm}p{0.3cm}p{0.3cm}|c|p{0.3cm}p{0.3cm}p{0.3cm}p{0.3cm}p{0.3cm}p{0.3cm}p{0.3cm}|c}
\toprule
\multirow{2}{*}{LLM} & \multicolumn{8}{c|}{\textbf{Chinese Commonsense Domain}} & \multicolumn{8}{c}{\textbf{Global Commonsense Domain}} \\
\cmidrule{2-17}
 & AJ & TU & SqU & MMR & SpU & NLI & RC & Avg. & AJ & TU & SqU & MMR & SpU & NLI & RC & Avg. \\
\midrule
Random Baseline & 50.0 & 25.0 & 25.0 &25.0 & 50.0 & 33.3 & 25.0 & 33.33 & 50.0 & 19.9 & 25.0 &25.0 & 50.0 & 33.3 & 25.0 & 32.60 \\
\midrule
GPT-3.5-1106 &  85.33 &   39.0 &    65.0 &    42.0 &    80.5 &    61.0 &   50.5 &          60.48 &  90.00 &   94.0 &    87.0 &    46.0 &    88.5 &    66.0 &   49.5 &          74.43 \\
GPT-4-1106 &  96.67 &   \textbf{60.0} &    85.0 &    74.0 &    86.0 &    77.0 &   62.5 &          \underline{77.31} &  \textbf{95.33} &   \textbf{98.0} &    \textbf{97.0 }&    \textbf{66.0 }&    90.0 &    72.0 &   \textbf{72.0} &          \textbf{84.33} \\
LLaMA-2-7B &  51.33 &   36.0 &    11.0 &    14.0 &    49.5 &    52.0 &    8.0 &          31.69 &  62.67 &   17.0 &    14.0 &    16.0 &    49.5 &    22.0 &   13.0 &          27.74 \\
LLaMA-2-13B  &  56.00 &   33.0 &    38.0 &    30.0 &    58.0 &    47.0 &   38.0 &          42.86 &  66.67 &   24.0 &    39.0 &    50.0 &    53.5 &    57.0 &   33.5 &          46.24 \\
LLaMA-2-70B &  57.33 &   37.0 &    52.0 &    32.0 &    55.0 &    56.0 &   41.5 &          47.26 &  72.67 &   84.0 &    73.0 &    42.0 &    64.0 &    61.0 &   41.5 &          62.60 \\
Vicuna-7B-v1.5 &  52.00 &   29.0 &    34.0 &    32.0 &    51.0 &    49.0 &   35.5 &          40.36 &  45.33 &   64.0 &    37.0 &    26.0 &    58.5 &    52.0 &   32.5 &          45.05 \\
Vicuna-13B-v1.5  &  64.67 &   25.0 &    32.0 &    26.0 &    51.5 &    60.0 &   40.0 &          42.74 &  72.67 &   74.0 &    41.0 &    50.0 &    68.0 &    61.0 &   36.0 &          57.52 \\
\midrule
ChatGLM3-6B &  66.00 &   40.0 &    59.0 &    38.0 &    77.0 &    72.0 &   37.5 &          55.64 &  34.00 &   69.0 &    71.0 &    28.0 &    75.5 &    63.0 &   34.0 &          53.50 \\
Baichuan2-7B  &  76.00 &   41.0 &    48.0 &    38.0 &    72.0 &    53.0 &   49.5 &          53.93 &  55.33 &   65.0 &    54.0 &    26.0 &    60.5 &    59.0 &   29.0 &          49.83 \\
Baichuan2-13B  &  85.33 &   40.0 &    48.0 &    46.0 &    72.5 &    66.0 &   51.5 &          58.48 &  77.33 &   74.0 &    58.0 &    40.0 &    71.0 &    61.0 &   39.0 &          60.05 \\
InternLM2-7B  &  88.00 &   38.0 &    58.0 &    38.0 &    76.0 &    81.0 &   25.0 &          57.71 &  74.67 &   80.0 &    62.0 &    20.0 &    78.0 &    \textbf{76.0} &   23.5 &          59.17 \\
InternLM2-20B  &  88.00 &   55.0 &    54.0 &    44.0 &    74.5 &    80.0 &   23.0 &          59.79 &  82.67 &   83.0 &    61.0 &    14.0 &    74.5 &    72.0 &   27.0 &          59.17 \\
Yi-6B &  70.67 &   32.0 &    47.0 &    32.0 &    75.0 &    50.0 &   42.0 &          49.81 &  79.33 &   63.0 &    43.0 &    14.0 &    70.5 &    57.0 &   33.5 &          51.48 \\
Yi-34B  &  96.00 &   55.0 &    89.0 &    76.0 &    \textbf{88.5} &    72.0 &   51.5 &          75.43 &  88.67 &   92.0 &    87.0 &    56.0 &    89.0 &    70.0 &   47.5 &          75.74 \\
DeepSeek-7B  &  81.33 &   34.0 &    50.0 &    50.0 &    79.5 &    57.0 &   31.5 &          54.76 &  68.00 &   76.0 &    47.0 &    50.0 &    72.5 &    59.0 &   32.5 &          57.86 \\
DeepSeek-67B  &  96.67 &   57.0 &    83.0 &    \textbf{92.0} &    87.5 &    77.0 &   34.5 &          75.38 &  90.00 &   95.0 &    86.0 &    22.0 &    88.0 &    73.0 &   39.0 &          70.43 \\
Qwen-7B  &  70.67 &   38.0 &    55.0 &    48.0 &    71.0 &    57.0 &   49.5 &          55.60 &  74.67 &   78.0 &    69.0 &    50.0 &    72.5 &    55.0 &   36.0 &          62.17 \\
Qwen-14B  &  87.33 &   54.0 &    77.0 &    60.0 &    82.5 &    66.0 &   55.0 &          68.83 &  84.00 &   83.0 &    83.0 &    44.0 &    84.5 &    71.0 &   40.0 &          69.93 \\
Qwen-72B  &  \textbf{98.00} &   59.0 &   \textbf{91.0} &    84.0 &    86.5 &    \textbf{84.0} &   \textbf{67.5} &          \textbf{81.43} &  94.00 &   92.0 &   93.0 &    64.0 &    \textbf{93.0} &    71.0 &   63.5 &          \underline{81.50} \\
\bottomrule
\end{tabular*}
\normalsize
    \caption{Accuracy on CHARM reasoning tasks. We selected the empirically optimal prompt strategy: XLT for English LLMs and ZH-CoT for Chinese-oriented LLMs. \textbf{Bold} and \underline{underline} represent the first and second place respectively. Detailed results are in Table \ref{tab:cn_tasks_reasoning_res_19LLMsx5p} and Table \ref{tab:gl_tasks_reasoning_res_19LLMsx5p} of Appendix \ref{sec:app_detailed_res_19x5}.}
    \label{tab:reasoning_res_19LLMs}
\end{table*}

\subsection{Prompt Strategies}

We selected 5 commonly used prompt strategies, and assessed the performance of the 19 LLMs on CHARM reasoning task:

\textbf{Direct}: The LLM does not perform intermediate reasoning and directly predicts the answer.

\textbf{ZH-CoT}: The LLM conducts intermediate reasoning \cite{wei2022cot} in Chinese before producing the answer.

\textbf{EN-CoT}: The reasoning process of CoT is in English for the 
 Chinese questions\cite{shi2022EN-cot}.

\textbf{Translate-EN}: We used the DeepL api\footnote{\url{https://www.deepl.com/translator}} to translate our benchmark into English, and then used English CoT for reasoning \cite{zhang2023donTrustChatGPT}.

\textbf{XLT}: The template prompt \cite{huang2023XLT} was used to change the original question into an English request, solve it step by step, and finally format the answer for output .

The examples for each prompt strategy are in Figure \ref{fig:examples-of-prompting} in Appendix 
 \ref{sec:app_examples_prompt}. For all prompt strategies, we use the 3-shot setting.

\section{Results and Analysis}
\label{sec:Results and Analysis}

\subsection{Integrated Reasoning Performance}
\label{sec:integrated-rea-perf}

We show the performance of the 19 LLMs on CHARM reasoning tasks in Table \ref{tab:reasoning_res_19LLMs}. We only choose one representative prompt strategy: XLT for English LLMs and ZH-CoT for Chinese-oriented LLMs, which is based on our empirical conclusion in \textsection \ref{sec:how_to_choose_prompt}. The LLMs' performance on the 7 aspects of the Chinese commonsense domain are shown in Table \ref{tab:cn_tasks_reasoning_7aspects_res_19LLMs} in Appendix \ref{sec:llm_perf_7aspects}.

\textbf{Commonsense Domain }  We found that the LLMs exhibit inconsistent performance in the global and Chinese commonsense domains. The rankings of the English LLMs dropped in the Chinese domain compared to the global domain. For instance, GPT-4 ranks first in the global domain, but in the Chinese domain, Qwen-72B outperforms all, pushing GPT-4 to the second. In the Chinese domain, the performance of LLaMA-2-70B is even worse than many Chinese-oriented LLMs in the 6B-7B size range.

However, in the global domain, LLaMA-2-70B is better than all Chinese-oriented LLMs up to 20B in size, except for Qwen-14B.

\subsection{Prompt Strategy Selection}
\label{sec:how_to_choose_prompt}

\begin{table}
    \centering
    \tiny 
\begin{tabular*}{0.49\textwidth}{llccc}
\toprule
         &\textbf{Prompt}& \textbf{Avg. all LLMs}&\textbf{Avg. CN-LLMs}&\textbf{Avg. EN-LLMs}\\
\midrule
Avg.  & Direct &          46.28 &    48.41 &              42.64 \\
all             & ZH-CoT &          56.66 &                       \textbf{62.40} &             46.81 \\
domains 
              & EN-CoT &          54.46 &                       58.19 &              48.06 \\
              & Translate-EN &          53.88 &                       55.51 &              51.07 \\
              & XLT &          56.81 &                       59.09 &              \textbf{52.90} \\
\midrule
Avg.  & Direct &          45.43 &                       47.76 &             41.44 \\
Chinese 
             & ZH-CoT &          \textbf{56.35} &                       \textbf{62.23} &              46.26 \\
domains 
              & EN-CoT &          52.06 &                       56.36 &              44.68 \\
              & Translate-EN &          47.25 &                       47.82 &              46.27 \\
              & XLT &          53.80 &                       56.63 &              \textbf{48.96} \\
\midrule
Avg. & Direct &          47.13 &                       49.05 &              43.85 \\
global              & ZH-CoT &          56.96 &                       62.57 &              47.35 \\
domains              & EN-CoT &          56.85 &                       60.01 &              51.44 \\
              & Translate-EN &          \textbf{60.50} &                       \textbf{63.20} &              55.87 \\
              & XLT &          59.82 &                       61.56 &              \textbf{56.84} \\
\bottomrule
\end{tabular*}
    \normalsize
    \caption{Averaged accuracy on CHARM reasoning tasks.  {\textquotedblleft CN-LLMs\textquotedblright}  means the 12 Chinese-oriented LLMs, {\textquotedblleft EN-LLMs\textquotedblright}  
 means the 7 English LLMs.}
    \label{tab:avg-along-two-dimension-integrated-rea}
\end{table}

We tested the combinations of the 19 LLMs and the 5 prompt strategies in CHARM reasoning tasks. Detailed results are in Table \ref{tab:cn_tasks_reasoning_res_19LLMsx5p} and Table \ref{tab:gl_tasks_reasoning_res_19LLMsx5p} in Appendix \ref{sec:app_detailed_res_19x5}. To draw some empirical conclusions, we analyzed along the following two dimensions:

\begin{dotpar}
Dim1: global or Chinese commonsense domain.
\end{dotpar}
\begin{dotpar}
Dim2: English or Chinese-oriented LLMs.
\end{dotpar}

We averaged the $19 \times 5$ LLM-prompt combinations along the above two dimensions, and the obtained results are in the Table \ref{tab:avg-along-two-dimension-integrated-rea}. 
\textit{From the LLM dimension}, it's clear that various LLMs prefer different prompt strategies: XLT consistently excels for English LLMs among the 5 strategies, while for Chinese-oriented LLMs, despite some complexity, ZH-CoT generally performs best. 
\textit{From the commonsense domain dimension}, strategies that use English for reasoning (like XLT, Translate-EN, etc.) are 
 suitable for the global domain; however, ZH-CoT generally performs better in the Chinese domain.

The conclusion here differs from previous studies \cite{shi2022EN-cot, huang2023XLT}, which suggested that employing English for non-English reasoning tasks was more effective than using the native language.
These previous studies had limitations, focusing only on English LLMs and neglecting the many Chinese-oriented LLMs developed since 2023. Furthermore, most benchmarks in these studies were merely translations from English, lacking unique cultural and linguistic characteristics in Chinese.
The empirical findings with CHARM in this paper have somewhat alleviated those limitations, leading to more current and comprehensive conclusions, and of course still have the limitations, which are detailed in section \textbf{Limitations}.

\subsection{Integrated Reasoning vs Memorization}
\label{sec:integrated-rea-vs-mem}

\begin{figure}
    \centering
    \includegraphics[width=1.0\linewidth]{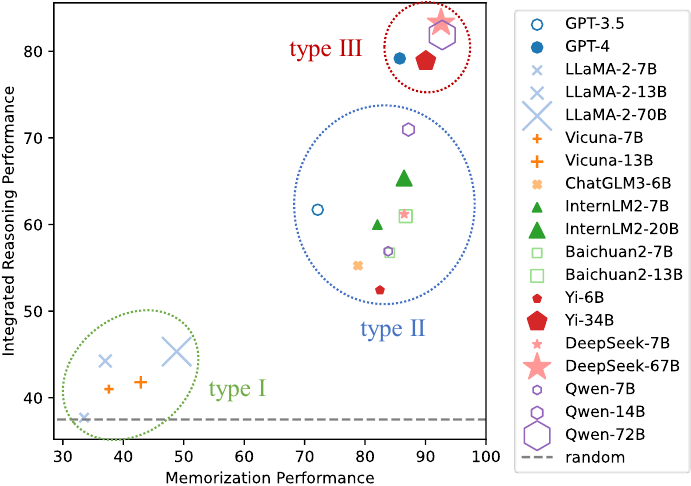}
    \caption{\textbf{Averaged} accuracy across the 4 \textit{MRI} tasks in the Chinese commonsense domain.}
    \label{fig:mem-integrated_rea-avg}
\end{figure}

We evaluated the correlation between the integrated reasoning and the memorization on the \textit{MRI} tasks, as mentioned in \textsection\ref{sec:reasoning_task_setup}. The average performance of the LLMs on the 4 \textit{MRI} tasks is in Figure \ref{fig:mem-integrated_rea-avg}. Detailed performance on each task is in Figure \ref{fig:mem-integrated_rea-4tasks} in Appendix \ref{sec:app_relation_mem_integrated_rea_4tasks}.

As shown in Figure \ref{fig:mem-integrated_rea-avg}, the 19 LLMs can be roughly divided into the three types:

\begin{dotpar}
\textbf{Type I: Low memorization and low integrated reasoning ability.} We found that apart from OpenAI's GPT series, all other English LLMs belong to this type.
\end{dotpar}
\begin{dotpar}
\textbf{Type II: High memorization and medium integrated reasoning ability.} GPT-3.5 and all Chinese-oriented LLMs below 30B belong to this type. It's worth noting that some LLMs have high memorization performance, but relatively poor integrated reasoning ability.
\end{dotpar}
\begin{dotpar}
\textbf{Type III: Ultra-high memorization and high integrated reasoning ability.} This category includes GPT-4 and the three Chinese-oriented LLMs that exceed a size of 30B.
\end{dotpar}

The above findings offer clear guidance for the enhancement of LLMs' reasoning abilities in Chinese commonsense domain. For Type I, the limitation lies in the memorization. For Type II, there should be further improvement in understanding, applying knowledge, and reasoning abilities.

In addition, we also evaluated the correlation between memorization and integrated reasoning during the LLM pre-training process, details can be found in Figure \ref{fig:mem-integrated_rea-pretraining-avg} in Appendix \ref{sec:app_relation_mem_integrated_rea_pretraining}.

The results clearly indicate that strong memorization is the foundation of integrated reasoning. Weak memorization leads to poor reasoning, as shown by Type I LLMs. Also, factors other than memorization can cause significant differences in reasoning abilities among LLMs with similar memorization.

\subsection{Memorization-Independent Reasoning}

\begin{table}
    \centering
\tiny
\begin{tabular*}{0.47\textwidth}{@{\extracolsep{\fill}} cccc}
\toprule
\multirow{2}{*}{\textbf{Rank}} & \multirow{2}{*}{\textbf{Integrated Reasoning}} & \multicolumn{2}{c}{\textbf{Memorization-independent Reasoning}}\\
\cmidrule{3-4}
& & \textbf{FRMM} & \textbf{MIB} \\
\midrule
1 & DeepSeek-67B & Yi-34B (\textcolor{red}{↑3}) & GPT-4-1106 (\textcolor{red}{↑2}) \\
2 & Qwen-72B & DeepSeek-67B (\textcolor{green}{↓1}) & Yi-34B (\textcolor{red}{↑2}) \\
3 & GPT-4-1106 & GPT-4-1106 (-) & Qwen-72B (\textcolor{green}{↓1}) \\
4 & Yi-34B & Qwen-72B (\textcolor{green}{↓2}) & DeepSeek-67B (\textcolor{green}{↓3}) \\
5 & Qwen-14B & GPT-3.5-1106 (\textcolor{red}{↑2}) & GPT-3.5-1106 (\textcolor{red}{↑2}) \\
6 & InternLM2-20B & Qwen-14B (\textcolor{green}{↓1}) & Qwen-14B (\textcolor{green}{↓1}) \\
7 & GPT-3.5-1106 & InternLM2-20B (\textcolor{green}{↓1}) & InternLM2-20B (\textcolor{green}{↓1}) \\
8 & InternLM2-7B & InternLM2-7B (-) & InternLM2-7B (-) \\
9 & DeepSeek-7B & Baichuan2-13B (\textcolor{red}{↑1}) & Baichuan2-13B (\textcolor{red}{↑1}) \\
10 & Baichuan2-13B & DeepSeek-7B (\textcolor{green}{↓1}) & DeepSeek-7B (\textcolor{green}{↓1}) \\
11 & Baichuan2-7B & Yi-6B (\textcolor{red}{↑3}) & Baichuan2-7B (-) \\
12 & ChatGLM3-6B& ChatGLM3-6B(-) & ChatGLM3-6B(-) \\
13 & Qwen-7B & Baichuan2-7B (\textcolor{green}{↓2}) & Qwen-7B (-) \\
14 & Yi-6B & Qwen-7B (\textcolor{green}{↓1}) & Yi-6B (-) \\
15 & LLaMA-2-70B & LLaMA-2-13B (\textcolor{red}{↑1}) & LLaMA-2-13B (\textcolor{red}{↑1}) \\
16 & LLaMA-2-13B & LLaMA-2-70B (\textcolor{green}{↓1}) & LLaMA-2-70B (\textcolor{green}{↓1}) \\
17 & Vicuna-13B-v1.5 & Vicuna-13B-v1.5 (-) & Vicuna-13B-v1.5 (-) \\
18 & Vicuna-7B-v1.5& LLaMA-2-7B (\textcolor{red}{↑1}) & Vicuna-7B-v1.5(-) \\
19 & LLaMA-2-7B & Vicuna-7B-v1.5 (\textcolor{green}{↓1}) & LLaMA-2-7B (-) \\
\bottomrule
\end{tabular*}
\normalsize
    \caption{Leaderboard on the \textit{MRI} tasks. We propose two methods, i.e. FRMM and MIC, to compare the LLMs' \textbf{memorization-independent reasoning}, as detailed in Appendix \ref{sec:app_leaderboard_mem_independent_rea}. The arrows and numbers in brackets in the last two columns indicate changes in ranking order relative to the second column.}
\label{tab:leaderboard-filter-rea-based-on-mem-19LLMs}
\end{table}

\textbf{Methods}
We proposed two methods, \textbf{FRMM} and \textbf{MIB}, to compare the LLMs' memorization-independent reasoning on the \textit{MRI} tasks. The results are in Table \ref{tab:leaderboard-filter-rea-based-on-mem-19LLMs}.

\begin{dotpar}
\textit{\textbf{F}iltering \textbf{R}easoning questions based on \textbf{M}ono-LLM-\textbf{M}emorization} (\textbf{FRMM})  For each LLM, we selected reasoning questions based on its performance in memorization tasks: only retaining reasoning questions for which all related memorization questions were answered correctly. Then we calculated the accuracy of the retained reasoning questions for each LLM. The LLMs were then ranked based on the accuracy, producing the leaderboard shown in the penultimate column of Table \ref{tab:leaderboard-filter-rea-based-on-mem-19LLMs}. The detail of the \textbf{FRMM} is in Appendix \ref{sec:app_filter_rea_mono}.
\end{dotpar}

\begin{dotpar}
\textit{\textbf{M}emorization-\textbf{I}ndependent \textbf{B}attles among LLMs} (\textbf{MIB}) Inspired by the pairwise battle method adopted in LLM evaluation \cite{zheng2023judging}, we tallyed each LLM's performance in a {\textquotedblleft round-robin\textquotedblright} tournament of pairwise match-ups and then ranked the performance of the LLMs.
Specifically, we selected two LLMs at each time and filter the \textit{MRI} task's reasoning questions based on the performance of these two LLMs in memorization tasks. We only retained the reasoning questions whose related memorization questions were correctly answered by both LLMs. In this way, the two LLMs were battled under fair conditions.
Then we calculated the accuracy of the two LLMs on the retained reasoning questions, and computed the difference in accuracy as the battle score between the two LLMs. For a total of 19 LLMs, we averaged each LLM's scores from the 18 battles they participated in as their final scores. The LLMs were then ranked based on these final scores, producing the leaderboard shown in the last column of Table \ref{tab:leaderboard-filter-rea-based-on-mem-19LLMs}.
The detail of the \textbf{MIB} is in Appendix \ref{sec:app_mem-independent-combats}. 
\end{dotpar}

\textbf{Analysis }
As shown in Table \ref{tab:leaderboard-filter-rea-based-on-mem-19LLMs}, when comparing the leaderboards for integrated and memorization-independent reasoning, Type III LLMs rank at the forefront and the Type I rank at the end in all leaderboards. There is a slight variation in the ranking order within the three types of LLMs.

\textbf{Error Types }
For the in-depth analysis, we chose Vicuna-13B, Qwen-7B, Qwen-72B as representatives for Type I, II, and III LLMs, and filtered out the reasoning questions in the \textit{MRI} tasks, only keeping those with correct answers to the related memorization questions, same as the FRMM in Appendix \ref{sec:app_filter_rea_mono}. This ensured the LLM had sufficiently memorized the commonsense knowledge required for the retained reasoning questions, thereby minimizing the impact of memorization on reasoning.
There are totally 500 reasoning questions in the 4 \textit{MRI} tasks, and the numbers of the retained are 106, 323 and 402 for Vicuna-13B, Qwen-7B and Qwen-72B respectively, as shown in Table \ref{tab:mem-based-filtering-of-rea}.

If the LLMs provided incorrect answers for the retained reasoning questions, these errors can be referred to as memorization-independent reasoning errors. We conducted the manual review and analysis of their reasoning process, and classified the errors into 4 main categories.

\begin{dotpar}
\textbf{Understanding Error } In this case, the LLM was unable to accurately comprehend the question, including misunderstanding the content, ignoring or even modifying important information in the premise, and failing to grasp the core query of the question.
\end{dotpar}
\begin{dotpar}
\textbf{Knowledge Error } The LLM incorporated inaccurate knowledge during the reasoning process. It's important to highlight that the knowledge pieces related to the reasoning question were previously examined in the related memorization questions, which the LLM answered \textbf{correctly}. However, the LLM output incorrect information during the reasoning phase.
\end{dotpar}
\begin{dotpar}
\textbf{Logical Error } The LLM made logical reasoning errors, such as mathematical reasoning errors, inability to reach the correct conclusion based on sufficient information, or reaching the correct conclusion but outputing the wrong option.
\end{dotpar}
\begin{dotpar}
\textbf{Other Errors } These are other scattered, relatively rare types of errors.
\end{dotpar}

\begin{figure}
    \centering
    \includegraphics[width=1.0\linewidth]{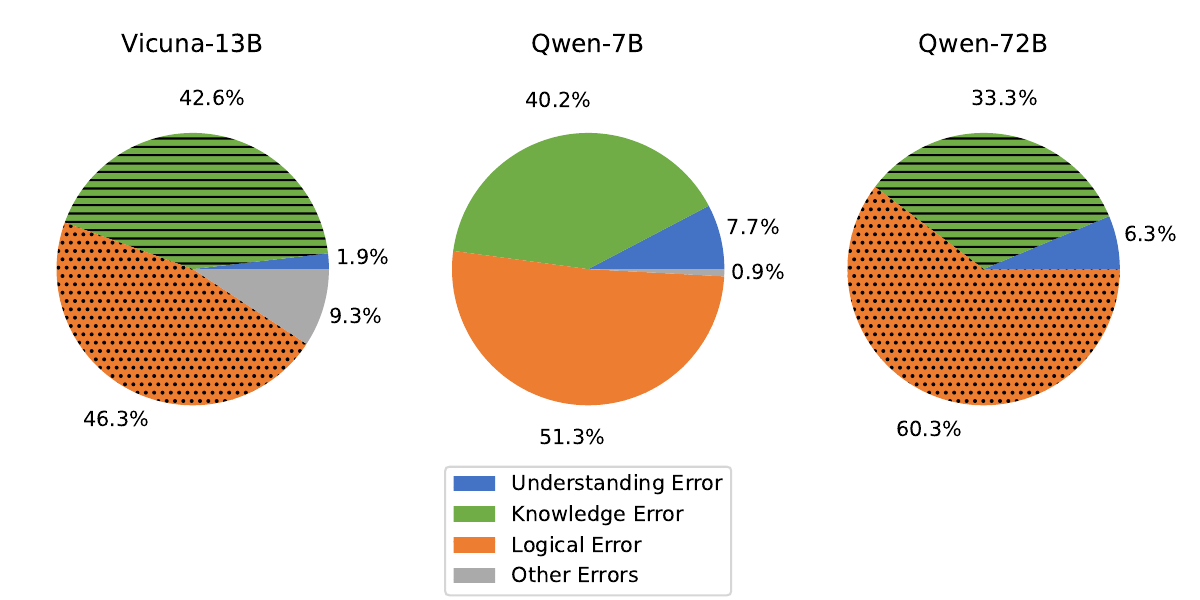}
    \caption{Distribution of the memorization-independent reasoning errors}
    \label{fig:pie-char-mem-independent-err-3LLMs}
\end{figure}

We show examples of each type of errors in Figure \ref{fig:mem-independent-rea-err-examples} in Appendix \ref{sec:app_mem_independent_rea_err}. The distribution of these error types are shown in Figure \ref{fig:pie-char-mem-independent-err-3LLMs}. 

\textbf{Discussions }
Obviously, the majority of errors are from logical reasoning mistakes and knowledge inaccuracies, which further provides the directions for LLMs' enhancement.  
As for knowledge errors, prior studies \cite{bian2023chatgptKnowledgeable,allen2023physics} have indicated that the way LLMs remember and master knowledge is a relatively complex topic. Simple memorization doesn't guarantee that LLMs can apply this knowledge accurately and skillfully during the reasoning process.

\begin{table}
    \centering
    \tiny
    \begin{tabular*}{0.48\textwidth}{@{\extracolsep{\fill}}cp{1cm}cccc}
    \toprule
\textbf{Models} & \textbf{LLM type} & \textbf{\# Original} & \textbf{\# Retained} & \textbf{\# Incorrect} \\
    \midrule
Vicuna-13B-v1.5 & Type I & 500 & 106 & 54 \\
Qwen-7B & Type II & 500 & 323 & 117 \\    
Qwen-72B & Type III & 500 & 402 & 63 \\    
    \bottomrule
    \end{tabular*}
    \normalsize
    \caption{Memorization-based filtering of reasoning questions. {\textquotedblleft Incorrect\textquotedblright} means the incorrectly answered questions among the \textbf{retained}.}
    \label{tab:mem-based-filtering-of-rea}
\end{table}

\section{Conclusion}
\label{sec:conclusion}
This paper introduces CHARM, the first benchmark designed to comprehensively and thoroughly evaluate LLMs' commonsense reasoning in Chinese. CHARM encompasses two counterpart commonsense domains, global and Chinese-specific, with the carefully selected tasks. We evaluated the representative prompt strategies for improving LLMs' reasoning ability, and the empirical findings significantly enhances and supplements the conclusions of previous studies. CHARM comprises closely-interconnected reasoning and memorization tasks, helping to reveal the intrinsic correlation between memorization and reasoning of LLMs. We evaluated the strengths and weaknesses of different LLMs and conducted the detailed analysis of memorization-independent reasoning abilities. We hope that CHARM's approach to studying the correlations between memoriztion and reasoning can serve as a reference for similar research in other fields.

\section*{Limitations}
\label{sec:limitations}

This study conducted tests on combinations of the 19 LLMs and the 5 prompt strategies, resulting in empirical conclusions. However, many existing LLMs and prompt strategies have not yet been tested. Furthermore, the best prompt strategy for the commonsense reasoning task for the LLMs, particularly in Chinese or other non-English languages, is not static and should progress with LLM technology. This is influenced by three elements: 
(1) The new prompt strategies are continuously proposed, which are likely more effective. 
(2) The new LLMs may have different prompt strategy preference, or be less sensitive to prompt.
(3) For other non-English languages with high resources, future LLMs would be continuously evolving and updating, and necessitate ongoing updates in evaluation.

The  automation of the construction and evaluation of CHARM needs further improvement, including the following:
(1) Most of the questions in CHARM Chinese domain are manually constructed by the author. This limits the number of benchmark questions and the range of knowledge pieces covered.
(2) Regarding memorization-independent reasoning, we chose only 3 LLMs as representative and manually categorized the types of errors within CHARM. In future research, we could  employ robust LLMs, like GPT-4, for automated error classification and statistical analysis.

\section*{Ethical Consideration}
This work involved human annotation. We have provided appropriate compensation for all annotators. The total cost of annotation for the project is about 2.2k RMB. For all annotators, we explicitly informed them about the use of the data and required them to ensure that the questions included in CHARM do not involve any social bias, ethical issues or privacy concerns during the annotation process.

\section*{Acknowledgements}
This research was supported by Shanghai Artificial Intelligence Laboratory. We thank Dr. Qipeng Guo of Shanghai AI Lab for his invaluable suggestions. We additionally thank our crowd annotators for their diligent work and the anonymous reviewers for their constructive comments.

\bibliography{ref}

\appendix

\section{Entity and Question Examples of the 7 Chinese Commonsense Aspects}
\begin{figure*}[!t]
    \centering
    \includegraphics[width=1\linewidth]{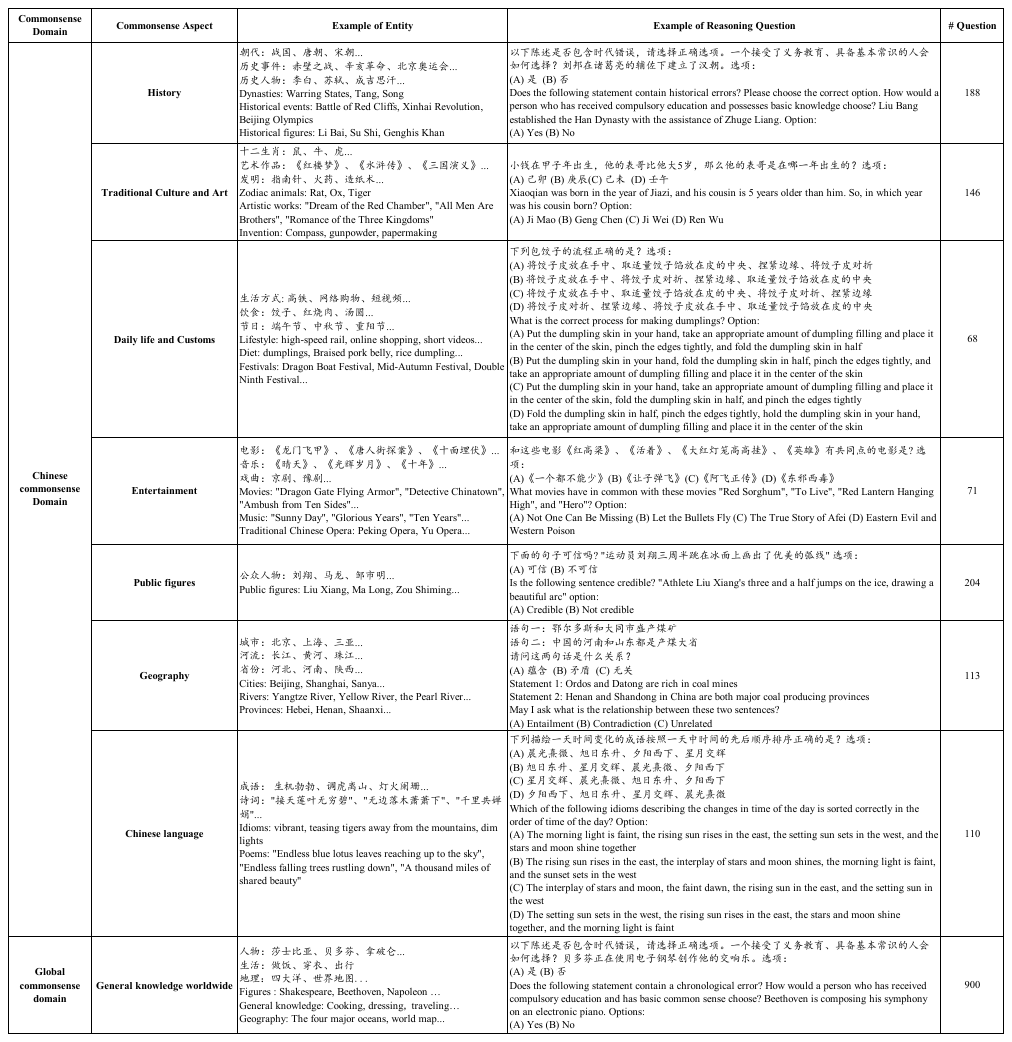}
    \caption{Entity and question examples of the commonsense aspects.}
    \label{fig:entity-and-question-examples-Chinese-commonsense-aspects}
\end{figure*}

Figure \ref{fig:entity-and-question-examples-Chinese-commonsense-aspects} shows the number of questions and partial entities of each Chinese commonsense aspect we propose, as well as corresponding question examples.

\section{Question Examples of the Reasoning Task in CHARM}
\label{sec:app_examples_rea_tasks}

\begin{figure*}
    \centering
    \includegraphics[width=1\linewidth]{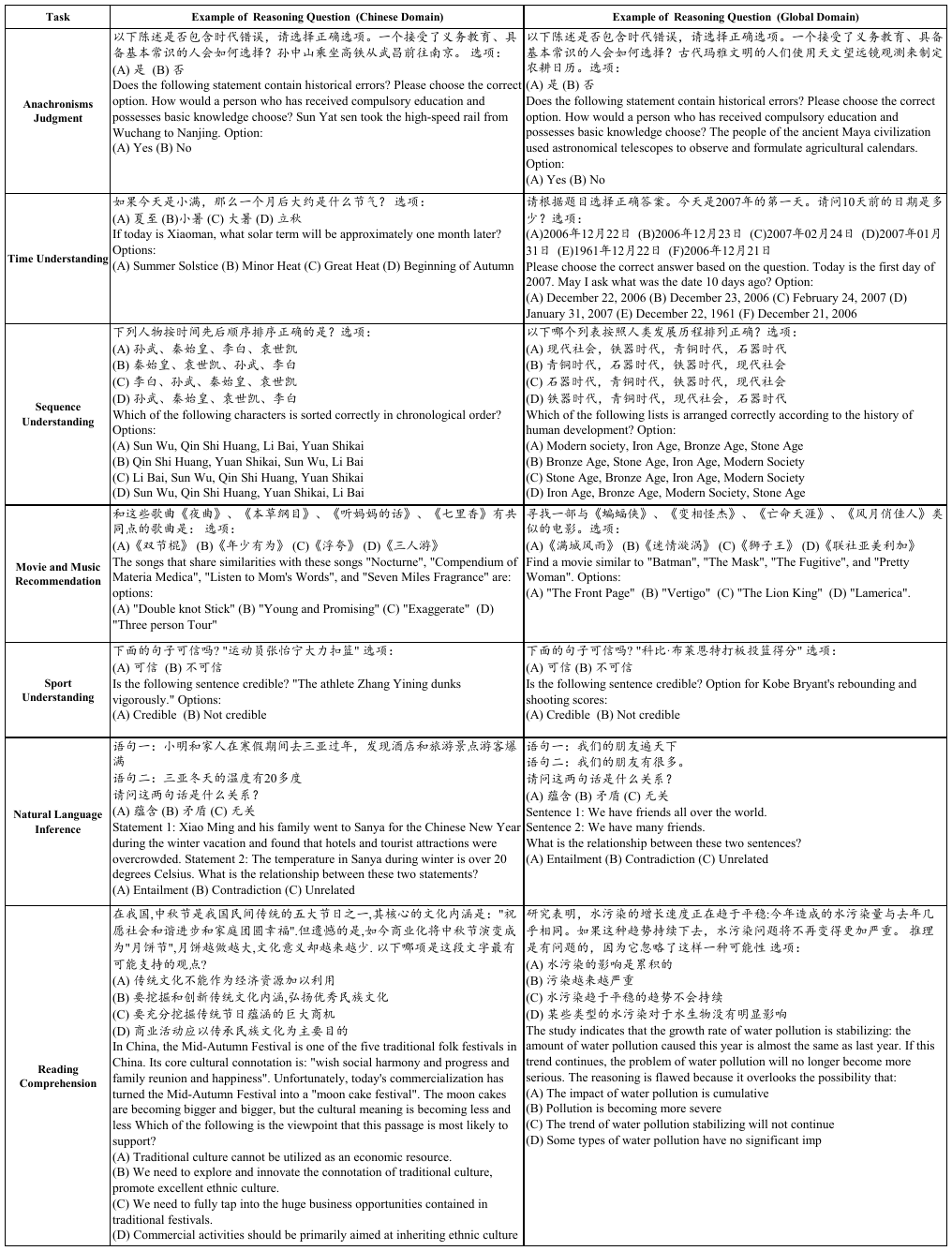}
    \caption{Examples of the reasoning tasks in CHARM.}
    \label{fig:task_rea_examples}
\end{figure*}

Figure \ref{fig:task_rea_examples} shows the question examples of the 7 reasoning tasks in CHARM, including both Chinese and global domains.

\section{Question Examples of the Memorization Task in CHARM}
\label{sec:app_examples_mem_tasks}

\begin{figure*}[!t]
    \centering
    \includegraphics[width=1\linewidth]{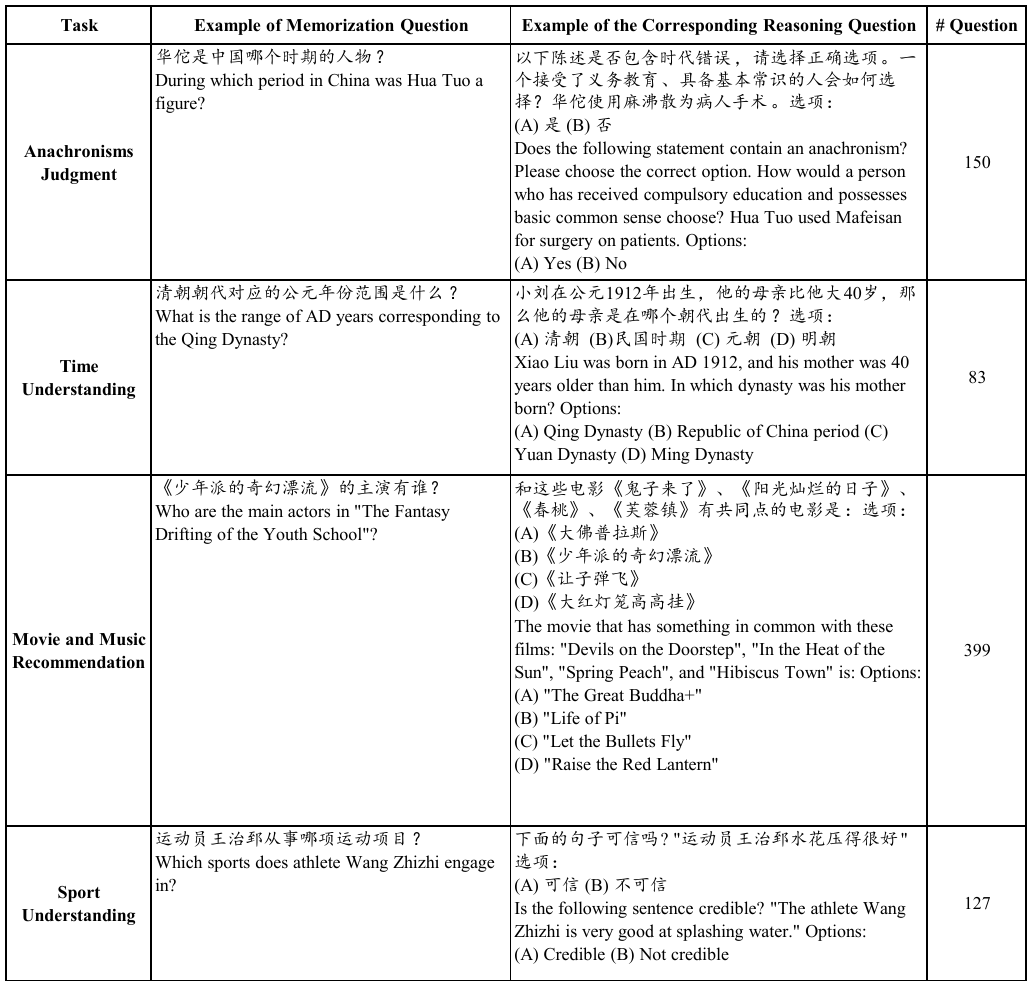}
    \caption{Examples of the memorization tasks in CHARM.}
    \label{fig:task_mem_example}
\end{figure*}

Figure \ref{fig:task_mem_example} shows the questions examples of the memorization tasks in CHARM.

\section{Details of Quality Assurance}
\label{sec:annotation_details}
\subsection{Quality Assurance Procedures}
After construction of CHARM, we conducted the quality assurance to ensure the quality of the questions. We hired professional NLP task annotators to review the questions we have constructed.
The entire quality assurance includes the following steps:

\textbf{Step(1) Annotation Task Submission and Assignment:} We packaged the constructed questions into annotation task packages. Usually, one annotation package corresponds to all the questions of one reasoning or memorization task in Table \ref{tab:charm_overview}. Before submitting the annotation tasks, we had written the annotation requirement document, which includes key requirements and typical examples. After the annotation tasks were submitted to the online annotation platform, they were assigned to suitable annotators. Typically, a task would be assigned to either two or four annotators.

\textbf{Step(2) Annotator training:} After the annotation task was assigned, we organized an online meeting to train the annotators. We specifically asked annotators to avoid social bias and sensitive issues such as morality.

\textbf{Step(3) Trial review: } Before the official review begun, we randomly selected 20\% of the questions for the annotators to try annotating. We reviewed the results of the trial annotation, corrected any detailed issues or understanding deviations in the review process in a timely manner, to ensure the quality of subsequent reviews.

\textbf{Step(4) Official review:} For each question, whether handcrafted or translated, we had two annotators do independent reviews. We only provided the question to the annotators, not our answers. Using external resources fully (such as search engines, online encyclopedias, etc.), the annotators would provide the answers. Annotators could also provide feedback on issues with the question itself (for example, translation errors or questions that do not meet the commonsense standards). Only when both annotators believed that there was no problem with the question, and their provided answers were consistent with our previous answers, did we consider the question to have passed the review.

\textbf{Step(5) Consultation and correction:} For questions that did not pass the step(4) review, several authors would hold a meeting to discuss the questions. There are three cases in total:

\begin{dotpar}
\textbf{Case(1) Retain:} There was no problem with the question itself, but the annotator answered incorrectly. We retained these questions.
\end{dotpar}
\begin{dotpar}
\textbf{Case(2) Discard:} The question had significant errors due to translation issues or problems with the question itself. We discarded these questions.
\end{dotpar}
\begin{dotpar}
\textbf{Case(3) Correct:} There were minor issues with the question or the answer, the multiple authors would discuss together and complete the correction of the questions.
\end{dotpar}

\subsection{Information of the Annotators}
We submitted the annotation task online to the professional data annotation company, which organized the annotators to complete the annotation work. A total of 30 professional annotators, all native Chinese speakers with extensive experience in natural language processing tasks, were involved in this project. They possess the expertise to discern and comprehend commonsense knowledge pertinent to both global and Chinese-specific contexts. Here is the specific information about these annotators.

\textbf{Education and profession:} Of the 30 annotators, 11 have a bachelor's degree and 19 have an associate degree. Regarding their fields of study, 9 are in humanities (4 specialized in design, 3 in business, 2 in language), while 21 are in science and engineering (11 specialized in computer science, 4 in automation, 3 in medicine, 2 in architecture, 1 in mathematics).

\textbf{Age:} 14 annotators are aged 20-25, 8 are aged 26-30, 7 are aged 30-35, and 1 is aged 35-40.

The annotators are all located in Changzhi City, Shanxi Province, China. We offer the hourly wage of 23.75 RMB for each annotator, which is higher than the local minimum wage standard. 

\subsection{Other Details}

During the construction of the CHARM project, we submitted a total of 22 annotation tasks, which together contained approximately 3.55k questions, and took a total of approximately 93 hours of the annotators' working time.
The entire process for each annotation task, from step(1) to step(4), typically required half a day. 
It needs to be emphasized that the annotators were only tasked with checking and answering questions that we have already created; \textit{they were NOT responsible for creating questions from scratch}.

The annotation work for the entire project spanned from January 2 to January 26, 2024. Each annotation task required only a few annotators to complete, rather than all 30 annotators. During the project period, the annotators participating in this project were also undertaking other data annotation tasks in the companies. In fact, our annotation tasks only accounted for a very small part of their working time.

The statistics results of the quality evaluation process are as followings: 

\begin{dotpar}
For reasoning tasks, the average failure rate in step(4) is 0.19, and the average ratio of case(3) in step(5) is 0.04. 
\end{dotpar}
\begin{dotpar}
For memorization tasks, the average failure rate in step(4) is 0.04, and the average ratio of case(3) in step(5) is 0.02.
\end{dotpar}

\section{Examples of Prompt Strategies}
\label{sec:app_examples_prompt}

\begin{figure*}[!t]
    \centering
    \includegraphics[width=1\linewidth]{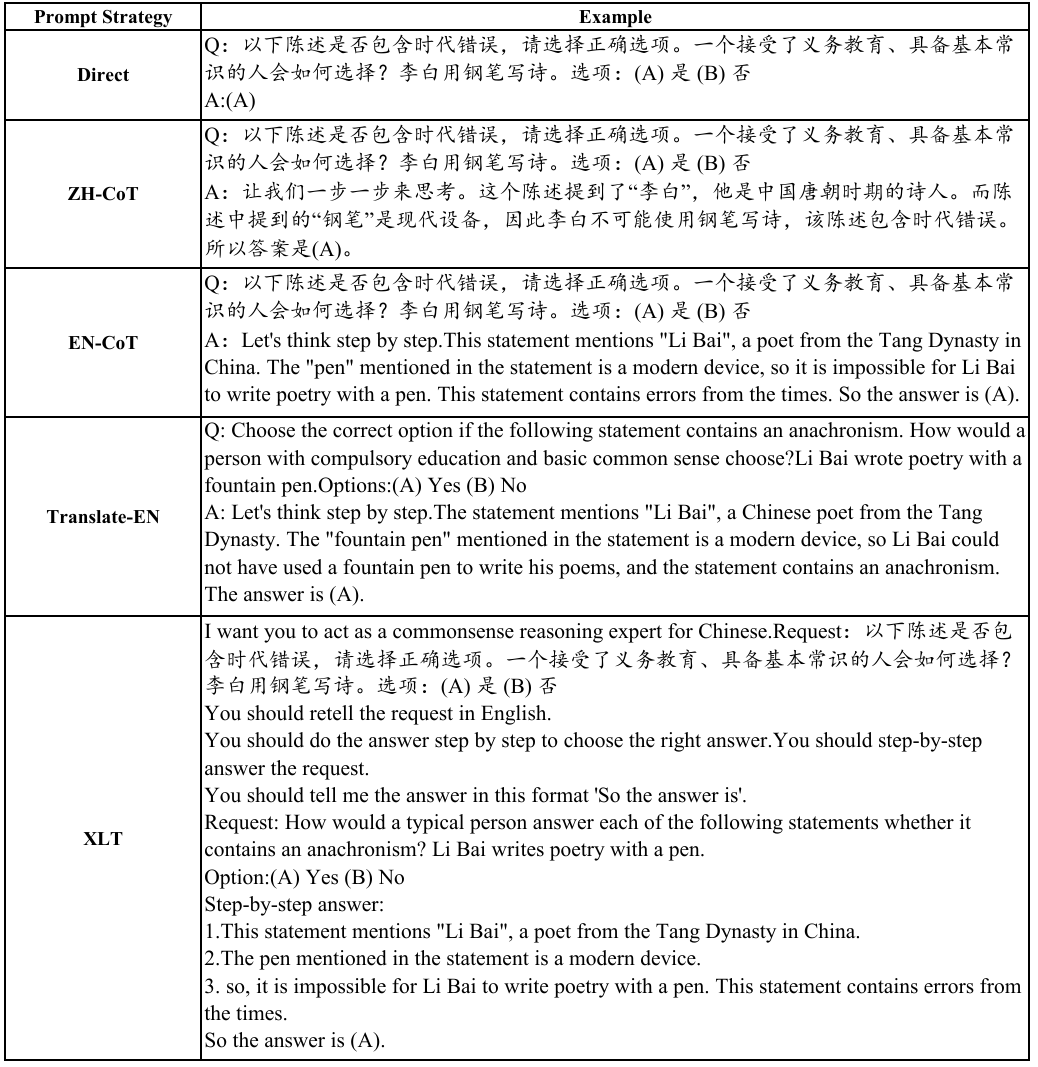}
    \caption{Examples of prompt strategies.}
    \label{fig:examples-of-prompting}
\end{figure*}

Figure \ref{fig:examples-of-prompting} shows the examples of the 5 prompt strategies.

\section{Detailed Evaluation Results of 19 LLMs with 5 Prompt Strageties on Reasoning Tasks}
\label{sec:app_detailed_res_19x5}
We conducted a detailed evaluation of 19 different LLMs using 5 distinct prompt strategies. Table \ref{tab:cn_tasks_reasoning_res_19LLMsx5p} and Table \ref{tab:gl_tasks_reasoning_res_19LLMsx5p} respectively display the performance of various prompt strategies on 7 reasoning tasks in the CHARM's Chinese commonsense domain and global commonsense domain.

\section{Performance of LLMs on Chinese Commonsense Knowledge Aspects}
\label{sec:llm_perf_7aspects}
Table \ref{tab:cn_tasks_reasoning_7aspects_res_19LLMs} displays the performance of LLMs in the 7 Chinese commonsense aspects. We only choose one representative prompt strategy: XLT for English LLMs and ZH-CoT for Chinese LLMs, which is based on our empirical conclusion in \textsection \ref{sec:how_to_choose_prompt}.

\section{Correlation of Memorization and Integrated Reasoning}
\label{sec:app_relation_mem_integrated_rea}

\subsection{Detailed Correlations of Memorization and Integrated Reasoning on the 4 \textit{MRI} Tasks}
\label{sec:app_relation_mem_integrated_rea_4tasks}

\begin{figure*}[!t]
    \centering
    \includegraphics[width=0.9\linewidth]{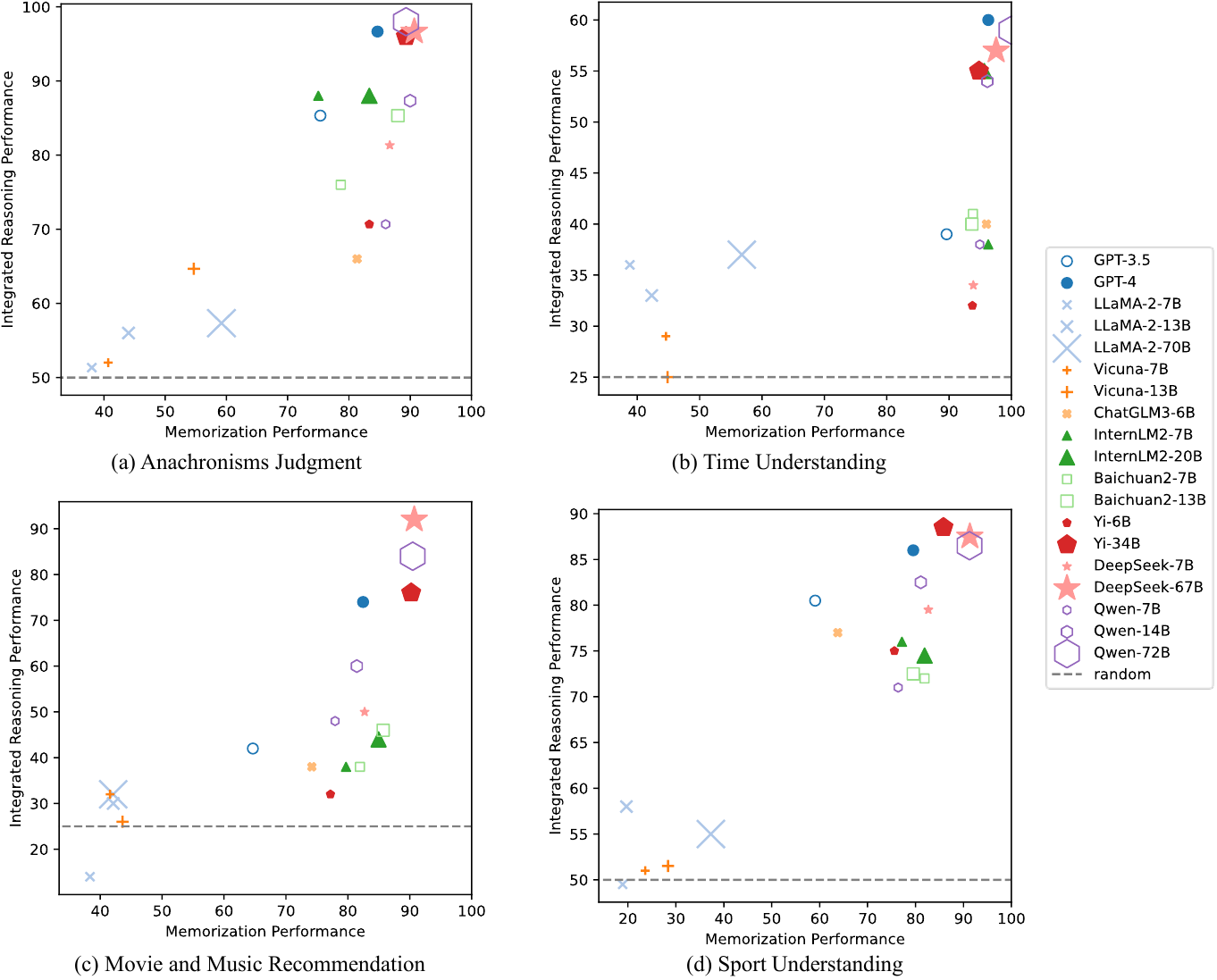}
    \caption{Accuracy of reasoning and memorization on the 4 \textit{MRI} tasks.}
    \label{fig:mem-integrated_rea-4tasks}
\end{figure*}

The detailed performances of the 19 LLMs on the 4 \textit{MRI} tasks are in Figure \ref{fig:mem-integrated_rea-4tasks}.

\subsection{Correlation of Memorization and Integrated Reasoning throughout the LLM pretraining}
\label{sec:app_relation_mem_integrated_rea_pretraining}

\begin{figure*}[!t]
    \centering
    \includegraphics[width=0.5\linewidth]{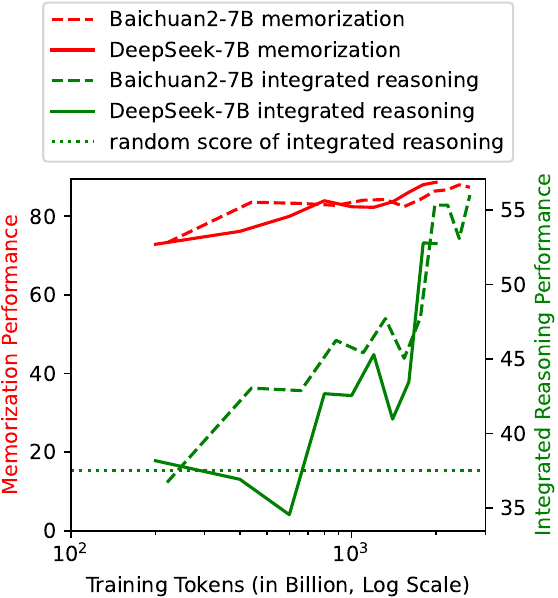}
    \caption{Averaged accuracy of the intermediate checkpoint models throughout the LLM pretraining across the 4 \textit{MRI} tasks.}
    \label{fig:mem-integrated_rea-pretraining-avg}
\end{figure*}

We tested the intermediate checkpoint models of Baichuan2 and DeepSeek on the memorization and reasoning questions on the 4 \textit{MRI} tasks. The results are shown in Figure \ref{fig:mem-integrated_rea-pretraining-avg}.

With the increase in the number of tokens during the training process, the model's memorization ability quickly reach a high level (in fact, there is no particularly obvious difference between the results of the first checkpoint and the final results). This is because the knowledge involved in our task setting is the most basic commonsense, and thus widely and abundantly exists in various Chinese training corpora.

However, the improvement in reasoning performance significantly lags behind memorization. This is because to complete a reasoning task in CHARM is actually a multi-step process, requiring memorization of relevant knowledge, understanding of the question, use of knowledge for reasoning, and answering according to the requirements of the question and the demonstration of few-shot examples, etc. If an error occurs in any step of the above complex process, the reasoning task will fail.

\section{Leaderboard of Memorization-Independent Reasoning}
\label{sec:app_leaderboard_mem_independent_rea}

It is non-trivial to acquire and compare the memorization-independent reasoning abilities of the LLMs. Intuitively, we can filter the reasoning questions by only retaining those whose related memorization questions are all correctedly answered by every LLMs. This approach ensures that each LLM has memorized the commonsense knowledge necessary for the retained reasoning questions. However, when we applied this process to all the 19 LLMs, only 28 reasoning questions remained out of the original 500 in the \textit{MRI} tasks, which was obviously insufficient in number and lacks diversity, thereby introducing a high degree of uncertainty due to randomness.

Therefore, we proposed two slightly more complex methods, one called \textit{Filtering Reasoning Questions based on Mono-LLM-Memorization} (FRMM), the other is \textit{Memorization-Independent Battles among LLMs} (MIB).

\subsection{Filtering Reasoning Questions based on Mono-LLM-Memorization (FRMM) }
\label{sec:app_filter_rea_mono}

This method is relatively simple, but has some flaws to a certain extent. For each LLM, we selected reasoning questions based on its performance in memorization tasks: only retaining reasoning questions for which all related memorization questions are answered correctly. It's clear that, after individual filtration, different LLMs would retain different reasoning questions, and even differ in the number of retained reasoning questions, as shown in the {\textquotedblleft\# retained\textquotedblright} column in Table \ref{tab:filter-rea-based-on-mem-19LLMs}.

Then, we calculated the accuracy of the retained reasoning questions for each LLM, and the results are shown in the {\textquotedblleft Retained Acc\textquotedblright} column in Table \ref{tab:filter-rea-based-on-mem-19LLMs}. The LLMs were then ranked based on the accuracy, producing the leaderboard shown in the penultimate column of Table \ref{tab:leaderboard-filter-rea-based-on-mem-19LLMs}.

As mentioned above, while this method can reflect the memorization-independent reasoning abilities of LLMs to some extent, its drawback lies in that the denominator used in calculating the final ranking accuracy differs for different LLMs.

To overcome this, we proposed the MIB method.

\subsection{Memorization-Independent Battles among LLMs (MIB) }
\label{sec:app_mem-independent-combats}

To overcome the shortcomings of the FRMM method, we referred to the pairwise battle method adopted in LLM evaluation \cite{zheng2023judging}. By tallying each LLM's performance in a {\textquotedblleft round-robin\textquotedblright} tournament of pairwise match-ups, we ranked the performance of the LLMs.

Specifically, we selected two LLMs at each time and filtered the \textit{MRI} task's reasoning questions based on the performance of these two LLMs in memorization tasks. We only retained the reasoning questions whose related memorization questions were correctly answered by both LLMs. In this way, the two LLMs are battled under fair conditions.
We then calculated the accuracy of these two LLMs on the retained reasoning questions separately, and compute the difference in accuracy as the battle score between the two models.

\begin{figure*}[!t]
    \centering
    \includegraphics[width=1.0\linewidth]{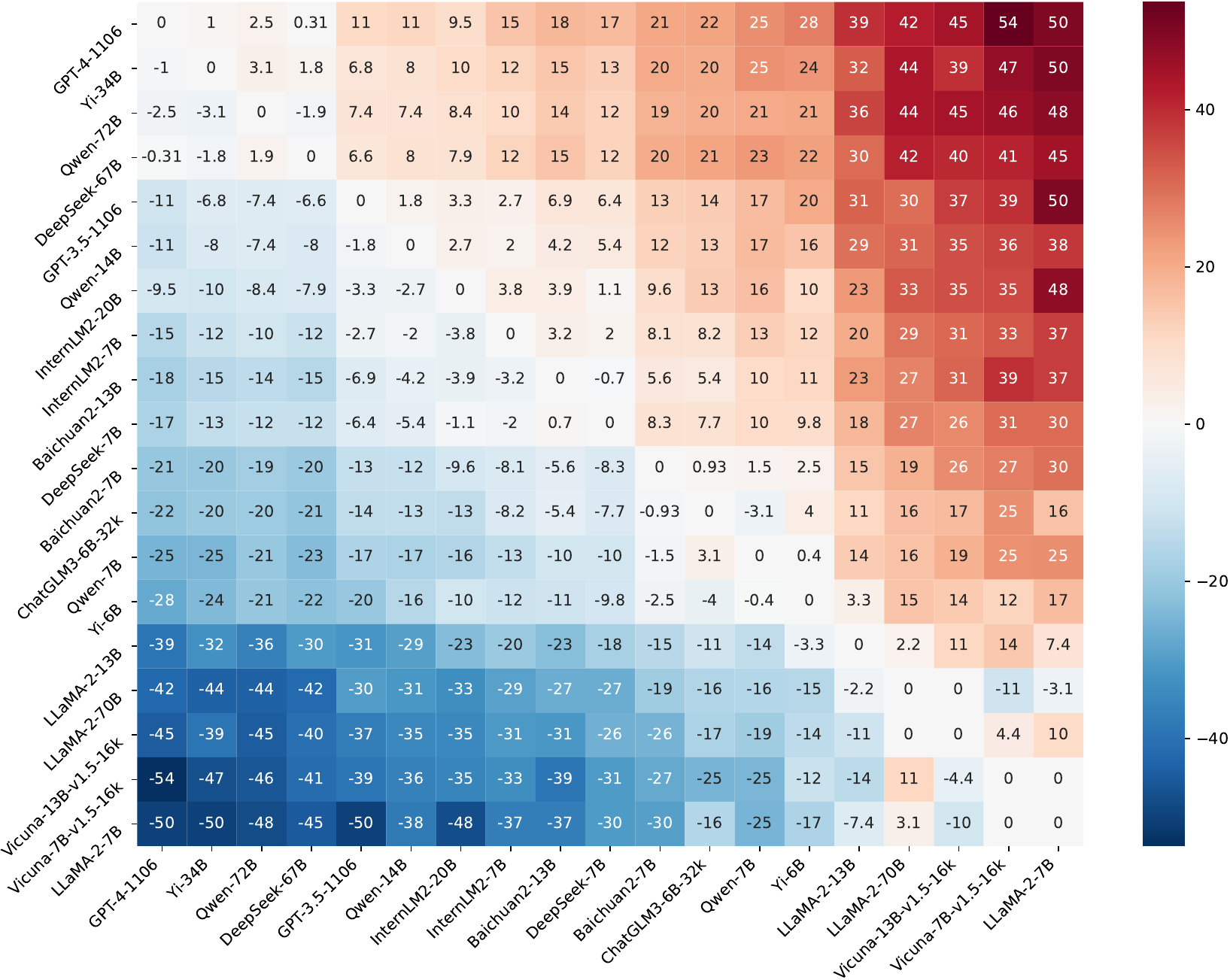}
    \caption{Results of the Memorization-Independent Battles among LLMs (MIB) on the \textit{MRI} tasks.}
    \label{fig:mem-independent-rea-combats}
\end{figure*}

As shown in Figure \ref{fig:mem-independent-rea-combats}, the element $E_{ij}$ represents the accuracy of LLM $i$ minus the accuracy of LLM $j$ during the battle between the two LLMs. For a total of 19 LLMs, we averaged each LLM's scores from the 18 battles they participated in as their final scores, as shown in Table \ref{tab:final-res-MIB-19LLMs}.

Finally, we ranked the LLMs based on these scores to produce the leaderboard shown in the last column of Table \ref{tab:leaderboard-filter-rea-based-on-mem-19LLMs}.

\section{Memorization-Independent Reasoning Errors}
\label{sec:app_mem_independent_rea_err}

\begin{figure*}[!t]
    \centering
    \includegraphics[width=1.0\linewidth]{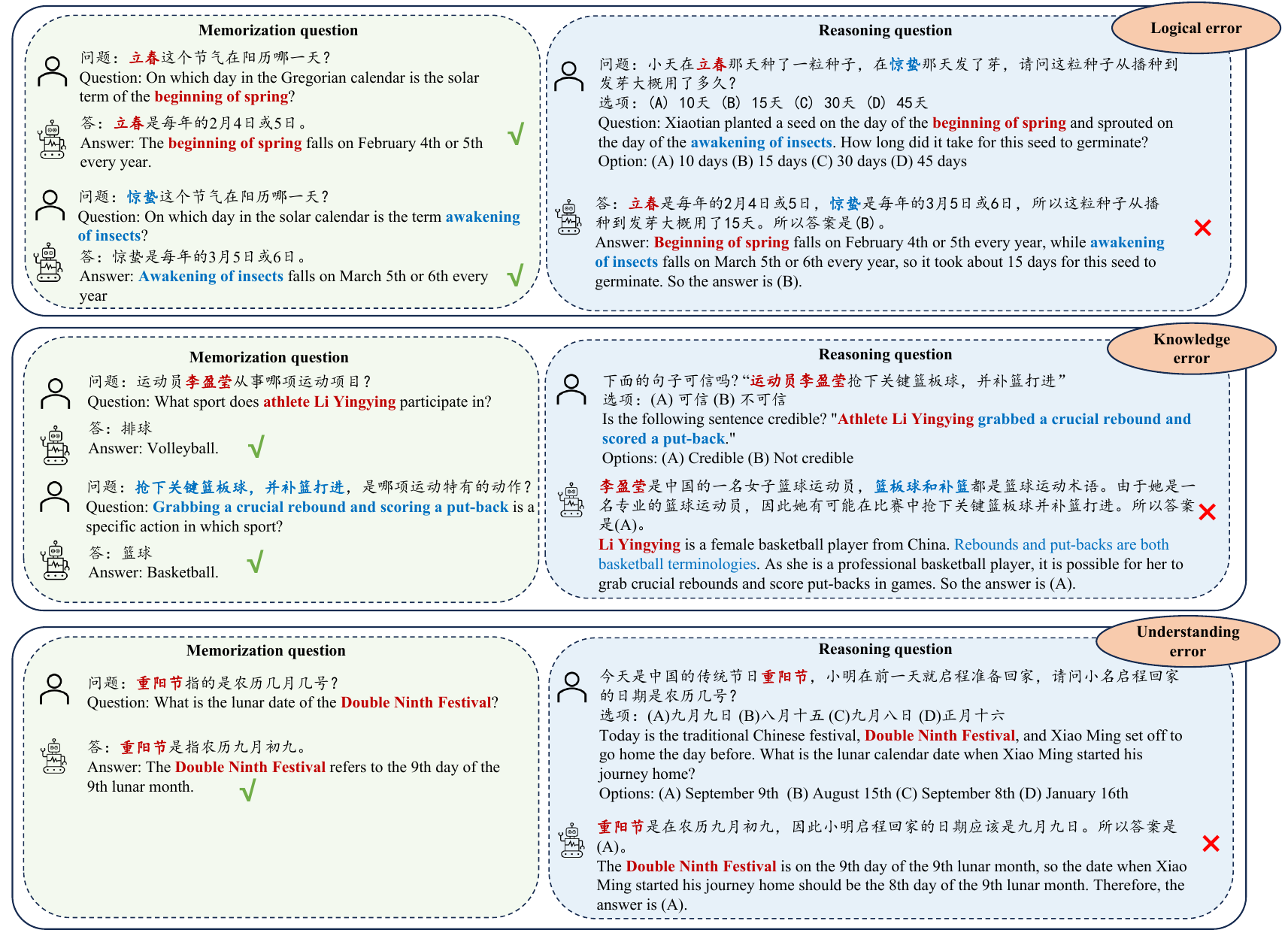}
    \caption{Examples of the 3 types of memorization-independent reasoning errors of LLMs}
    \label{fig:mem-independent-rea-err-examples}
\end{figure*}

LLMs can answer memorization questions correctly, but they make mistakes when it comes to reasoning problems composed of these knowledge points. Figure \ref{fig:mem-independent-rea-err-examples} shows the examples of three memorization-independent reasoning errors of LLMs.

\begin{table*}
    \centering
\tiny
\begin{tabular}{llrrrrrrrr}
\toprule
LLM & Prompt &  AJ &  TU &  SqU &  MMR &  SpU &  NLI &  RC &  Avg. \\
\midrule
GPT-3.5-1106 & Direct &  32.00 &   33.0 &    46.0 &    48.0 &    60.0 &    62.0 &   52.5 &          47.64 \\
         & ZH-CoT &  17.33 &   49.0 &    69.0 &    46.0 &    69.0 &    66.0 &   55.0 &          53.05 \\
         & EN-CoT &  50.67 &   39.0 &    64.0 &    44.0 &    80.5 &    59.0 &   53.5 &          55.81 \\
         & Translate-EN &  87.33 &   44.0 &    54.0 &    40.0 &    76.5 &    64.0 &   49.5 &          59.33 \\
         & XLT &  85.33 &   39.0 &    65.0 &    42.0 &    80.5 &    61.0 &   50.5 &          60.48 \\
\hline
GPT-4-1106 & Direct &  94.00 &   50.0 &    81.0 &    72.0 &    73.0 &    86.0 &   63.0 &          74.14 \\
         & ZH-CoT &  99.33 &   67.0 &    81.0 &    68.0 &    85.0 &    83.0 &   34.0 &          73.90 \\
         & EN-CoT &  98.00 &   65.0 &    79.0 &    70.0 &    83.0 &    83.0 &   50.5 &          75.50 \\
         & Translate-EN &  96.00 &   43.0 &    70.0 &    42.0 &    75.5 &    81.0 &   53.5 &          65.86 \\
         & XLT &  96.67 &   60.0 &    85.0 &    74.0 &    86.0 &    77.0 &   62.5 &          77.31 \\
         \hline
LLaMA-2-7B & Direct &  47.33 &   21.0 &    20.0 &    30.0 &    49.0 &     6.0 &   31.5 &          29.26 \\
         & ZH-CoT &  52.67 &   34.0 &    23.0 &    22.0 &    51.0 &    34.0 &   29.5 &          35.17 \\
         & EN-CoT &  44.67 &   34.0 &    20.0 &     4.0 &    50.5 &    40.0 &   26.5 &          31.38 \\
         & Translate-EN &  10.00 &   32.0 &    20.0 &    20.0 &    39.0 &    40.0 &   20.0 &          25.86 \\
         & XLT &  51.33 &   36.0 &    11.0 &    14.0 &    49.5 &    52.0 &    8.0 &          31.69 \\
         \hline
LLaMA-2-13B & Direct &  47.33 &   30.0 &    34.0 &    30.0 &    49.0 &    33.0 &   31.5 &          36.40 \\
         & ZH-CoT &  52.67 &   38.0 &    38.0 &    30.0 &    53.0 &    34.0 &   23.0 &          38.38 \\
         & EN-CoT &  52.67 &   34.0 &    38.0 &     2.0 &    49.0 &     1.0 &   35.5 &          30.31 \\
         & Translate-EN &  53.33 &   34.0 &    20.0 &    10.0 &    62.0 &     9.0 &   31.0 &          31.33 \\
         & XLT &  56.00 &   33.0 &    38.0 &    30.0 &    58.0 &    47.0 &   38.0 &          42.86 \\
         \hline
LLaMA-2-70B & Direct &  47.33 &   23.0 &    25.0 &    26.0 &    49.5 &    44.0 &   33.0 &          35.40 \\
         & ZH-CoT &  51.33 &   31.0 &    35.0 &    24.0 &    51.5 &    46.0 &   37.0 &          39.40 \\
         & EN-CoT &  55.33 &   27.0 &    31.0 &    26.0 &    59.0 &    56.0 &   41.0 &          42.19 \\
         & Translate-EN &  72.67 &   26.0 &    46.0 &    42.0 &    66.5 &    65.0 &   48.0 &          52.31 \\
         & XLT &  57.33 &   37.0 &    52.0 &    32.0 &    55.0 &    56.0 &   41.5 &          47.26 \\
         \hline
Vicuna-7B-v1.5 & Direct &  52.67 &   25.0 &    30.0 &    16.0 &    14.5 &    19.0 &   21.5 &          25.52 \\
         & ZH-CoT &  56.00 &   25.0 &    39.0 &    26.0 &    49.5 &    56.0 &   33.0 &          40.64 \\
         & EN-CoT &  53.33 &   28.0 &    26.0 &    18.0 &    40.5 &    55.0 &   40.5 &          37.33 \\
         & Translate-EN &  66.67 &   25.0 &    31.0 &    30.0 &    60.0 &    57.0 &   33.0 &          43.24 \\
         & XLT &  52.00 &   29.0 &    34.0 &    32.0 &    51.0 &    49.0 &   35.5 &          40.36 \\
         \hline
Vicuna-13B-v1.5 & Direct &  47.33 &   34.0 &    34.0 &    30.0 &    48.5 &    52.0 &   46.0 &          41.69 \\
         & ZH-CoT &  64.00 &   33.0 &    31.0 &    34.0 &    51.0 &    54.0 &   36.0 &          43.29 \\
         & EN-CoT &  62.67 &   30.0 &    32.0 &    24.0 &    50.0 &    50.0 &   33.0 &          40.24 \\
         & Translate-EN &  69.33 &   23.0 &    26.0 &    32.0 &    63.0 &    68.0 &   40.5 &          45.98 \\
         & XLT &  64.67 &   25.0 &    32.0 &    26.0 &    51.5 &    60.0 &   40.0 &          42.74 \\
         \hline
ChatGLM3-6B & Direct &  44.67 &   35.0 &    48.0 &    46.0 &    58.5 &    73.0 &   60.5 &          52.24 \\
         & ZH-CoT &  66.00 &   40.0 &    59.0 &    38.0 &    77.0 &    72.0 &   37.5 &          55.64 \\
         & EN-CoT &  55.33 &   36.0 &    58.0 &    36.0 &    76.0 &    75.0 &   32.5 &          52.69 \\
         & Translate-EN &  57.33 &   24.0 &    39.0 &    40.0 &    53.5 &    71.0 &   46.0 &          47.26 \\
         & XLT &  48.00 &   44.0 &    43.0 &    36.0 &    68.5 &    65.0 &   41.0 &          49.36 \\
         \hline
Baichuan2-7B & Direct &  44.67 &   31.0 &    37.0 &    24.0 &    59.0 &    35.0 &   56.0 &          40.95 \\
         & ZH-CoT &  76.00 &   41.0 &    48.0 &    38.0 &    72.0 &    53.0 &   49.5 &          53.93 \\
         & EN-CoT &  55.33 &   36.0 &    44.0 &    30.0 &    69.0 &    53.0 &   41.0 &          46.90 \\
         & Translate-EN &  55.33 &   21.0 &    26.0 &    24.0 &    41.5 &    52.0 &   33.5 &          36.19 \\
         & XLT &  56.00 &   35.0 &    44.0 &    28.0 &    68.0 &    48.0 &   44.0 &          46.14 \\
         \hline
Baichuan2-13B & Direct &  59.33 &   23.0 &    42.0 &    30.0 &    67.0 &    36.0 &   23.5 &          40.12 \\
         & ZH-CoT &  85.33 &   40.0 &    48.0 &    46.0 &    72.5 &    66.0 &   51.5 &          58.48 \\
         & EN-CoT &  72.00 &   40.0 &    50.0 &    34.0 &    68.0 &    64.0 &   42.5 &          52.93 \\
         & Translate-EN &  65.33 &   38.0 &    40.0 &    32.0 &    58.5 &    49.0 &   36.0 &          45.55 \\
         & XLT &  61.33 &   33.0 &    38.0 &    34.0 &    67.0 &    61.0 &   46.0 &          48.62 \\
         \hline
InternLM2-7B & Direct &  22.00 &   33.0 &    62.0 &    54.0 &    58.5 &    84.0 &   66.0 &          54.21 \\
         & ZH-CoT &  88.00 &   38.0 &    58.0 &    38.0 &    76.0 &    81.0 &   25.0 &          57.71 \\
         & EN-CoT &  77.33 &   42.0 &    59.0 &    38.0 &    73.0 &    78.0 &   38.5 &          57.98 \\
         & Translate-EN &  73.33 &   28.0 &    45.0 &    36.0 &    56.5 &    72.0 &   46.0 &          50.98 \\
         & XLT &  80.67 &   38.0 &    60.0 &    30.0 &    66.5 &    72.0 &   53.5 &          57.24 \\
         \hline
InternLM2-20B & Direct &  14.00 &   42.0 &    61.0 &    50.0 &    39.5 &    54.0 &   46.5 &          43.86 \\
         & ZH-CoT &  88.00 &   55.0 &    54.0 &    44.0 &    74.5 &    80.0 &   23.0 &          59.79 \\
         & EN-CoT &  68.67 &   40.0 &    48.0 &    42.0 &    67.0 &    68.0 &   25.0 &          51.24 \\
         & Translate-EN &  80.67 &   34.0 &    54.0 &    36.0 &    53.5 &    71.0 &   53.0 &          54.60 \\
         & XLT &  85.33 &   36.0 &    71.0 &    42.0 &    64.5 &    68.0 &   58.0 &          60.69 \\
         \hline
Yi-6B & Direct &  14.67 &   17.0 &    20.0 &    30.0 &    48.0 &    19.0 &   35.5 &          26.31 \\
         & ZH-CoT &  70.67 &   32.0 &    47.0 &    32.0 &    75.0 &    50.0 &   42.0 &          49.81 \\
         & EN-CoT &  58.67 &   18.0 &    34.0 &    30.0 &    58.0 &    52.0 &   48.5 &          42.74 \\
         & Translate-EN &  56.00 &   25.0 &    23.0 &    26.0 &    24.0 &    15.0 &   23.0 &          27.43 \\
         & XLT &  54.67 &   36.0 &    35.0 &    28.0 &    68.5 &    56.0 &   43.0 &          45.88 \\
         \hline
Yi-34B & Direct &  89.33 &   28.0 &    85.0 &    56.0 &    70.0 &    51.0 &   68.0 &          63.90 \\
         & ZH-CoT &  96.00 &   55.0 &    89.0 &    76.0 &    88.5 &    72.0 &   51.5 &          75.43 \\
         & EN-CoT &  90.00 &   42.0 &    78.0 &    66.0 &    84.5 &    67.0 &   50.0 &          68.21 \\
         & Translate-EN &  86.67 &   28.0 &    55.0 &    34.0 &    71.0 &    65.0 &   41.5 &          54.45 \\
         & XLT &  92.00 &   48.0 &    87.0 &    72.0 &    84.0 &    66.0 &   61.0 &          72.86 \\
         \hline
DeepSeek-7B & Direct &  46.67 &   27.0 &    21.0 &    30.0 &    48.0 &    40.0 &   27.5 &          34.31 \\
         & ZH-CoT &  81.33 &   34.0 &    50.0 &    50.0 &    79.5 &    57.0 &   31.5 &          54.76 \\
         & EN-CoT &  72.00 &   33.0 &    33.0 &    24.0 &    73.0 &    47.0 &   35.5 &          45.36 \\
         & Translate-EN &  68.67 &   18.0 &    28.0 &    36.0 &    59.0 &    72.0 &   40.5 &          46.02 \\
         & XLT &  55.33 &   32.0 &    39.0 &    36.0 &    51.0 &    37.0 &   35.0 &          40.76 \\
         \hline
DeepSeek-67B & Direct &  22.67 &   48.0 &    33.0 &    28.0 &    12.5 &    53.0 &   39.5 &          33.81 \\
         & ZH-CoT &  96.67 &   57.0 &    83.0 &    92.0 &    87.5 &    77.0 &   34.5 &          75.38 \\
         & EN-CoT &  84.00 &   53.0 &    73.0 &    58.0 &    82.0 &    73.0 &   35.0 &          65.43 \\
         & Translate-EN &  94.67 &   45.0 &    60.0 &    38.0 &    67.5 &    69.0 &   40.0 &          59.17 \\
         & XLT &  95.33 &   59.0 &    80.0 &    66.0 &    87.5 &    76.0 &   54.5 &          74.05 \\
         \hline
Qwen-7B & Direct &  50.67 &   28.0 &    41.0 &    50.0 &    60.5 &    56.0 &   55.5 &          48.81 \\
         & ZH-CoT &  70.67 &   38.0 &    55.0 &    48.0 &    71.0 &    57.0 &   49.5 &          55.60 \\
         & EN-CoT &  58.67 &   40.0 &    48.0 &    32.0 &    68.5 &    58.0 &   43.0 &          49.74 \\
         & Translate-EN &  62.67 &   23.0 &    31.0 &    26.0 &    60.0 &    54.0 &   45.0 &          43.10 \\
         & XLT &  62.67 &   26.0 &    47.0 &    40.0 &    63.0 &    50.0 &   50.5 &          48.45 \\
         \hline
Qwen-14B & Direct &  63.33 &   28.0 &    69.0 &    60.0 &    73.0 &    59.0 &   59.0 &          58.76 \\
         & ZH-CoT &  87.33 &   54.0 &    77.0 &    60.0 &    82.5 &    66.0 &   55.0 &          68.83 \\
         & EN-CoT &  85.33 &   48.0 &    68.0 &    56.0 &    77.5 &    76.0 &   53.0 &          66.26 \\
         & Translate-EN &  78.00 &   29.0 &    45.0 &    22.0 &    60.5 &    63.0 &   46.5 &          49.14 \\
         & XLT &  83.33 &   36.0 &    66.0 &    56.0 &    76.5 &    65.0 &   50.0 &          61.83 \\
         \hline
Qwen-72B & Direct &  90.67 &   36.0 &    85.0 &    78.0 &    80.0 &    84.0 &   77.5 &          75.88 \\
         & ZH-CoT &  98.00 &   59.0 &    91.0 &    84.0 &    86.5 &    84.0 &   67.5 &          81.43 \\
         & EN-CoT &  95.33 &   55.0 &    88.0 &    64.0 &    86.0 &    78.0 &   72.0 &          76.90 \\
         & Translate-EN &  92.00 &   37.0 &    53.0 &    32.0 &    73.0 &    76.0 &   57.0 &          60.00 \\
         & XLT &  93.33 &   50.0 &    86.0 &    70.0 &    83.0 &    75.0 &   58.5 &          73.69 \\
\bottomrule
\end{tabular}
\normalsize
    \caption{Accuracy of reasoning tasks in the \textbf{Chinese} commonsense domain of CHARM.}
    \label{tab:cn_tasks_reasoning_res_19LLMsx5p}
\end{table*}

\begin{table*}
    \centering
\tiny
\begin{tabular}{llrrrrrrrr}
\toprule
LLM & Prompt &  AJ &  TU &  SqU &  MMR &  SpU &  NLI &  RC &  Avg. \\
\midrule
GPT-3.5-1106 & Direct &  41.33 &   58.0 &    59.0 &    42.0 &    61.0 &    64.0 &   45.0 &          52.90 \\
         & ZH-CoT &  25.33 &   89.0 &    90.0 &    28.0 &    77.5 &    73.0 &   48.5 &          61.62 \\
         & EN-CoT &  59.33 &   85.0 &    84.0 &    42.0 &    86.0 &    68.0 &   50.0 &          67.76 \\
         & Translate-EN &  88.67 &   86.0 &    80.0 &    48.0 &    83.5 &    65.0 &   58.0 &          72.74 \\
         & XLT &  90.00 &   94.0 &    87.0 &    46.0 &    88.5 &    66.0 &   49.5 &          74.43 \\
\hline
GPT-4-1106 & Direct &  90.67 &   83.0 &    92.0 &    70.0 &    88.0 &    78.0 &   74.0 &          82.24 \\
         & ZH-CoT &  92.67 &  100.0 &    90.0 &    34.0 &    88.0 &    76.0 &   50.5 &          75.88 \\
         & EN-CoT &  95.33 &   97.0 &    97.0 &    52.0 &    89.5 &    70.0 &   61.5 &          80.33 \\
         & Translate-EN &  92.67 &   93.0 &    91.0 &    48.0 &    74.0 &    68.0 &   62.0 &          75.52 \\
         & XLT &  95.33 &   98.0 &    97.0 &    66.0 &    90.0 &    72.0 &   72.0 &          84.33 \\
\hline
LLaMA-2-7B & Direct &  43.33 &   20.0 &    20.0 &    28.0 &    51.5 &    18.0 &   20.5 &          28.76 \\
         & ZH-CoT &  51.33 &   18.0 &    22.0 &    26.0 &    50.0 &    31.0 &   22.5 &          31.55 \\
         & EN-CoT &  56.67 &   20.0 &    22.0 &    22.0 &    51.5 &    10.0 &   27.0 &          29.88 \\
         & Translate-EN &   4.67 &   20.0 &    35.0 &     2.0 &    51.5 &    38.0 &   28.0 &          25.60 \\
         & XLT &  62.67 &   17.0 &    14.0 &    16.0 &    49.5 &    22.0 &   13.0 &          27.74 \\
\hline
LLaMA-2-13B & Direct &  53.33 &   21.0 &    35.0 &    24.0 &    51.0 &    35.0 &   20.5 &          34.26 \\
         & ZH-CoT &  54.67 &   15.0 &    35.0 &    14.0 &    51.5 &    32.0 &   27.0 &          32.74 \\
         & EN-CoT &  52.67 &   19.0 &    35.0 &    16.0 &    51.5 &    35.0 &   38.0 &          35.31 \\
         & Translate-EN &  34.67 &   19.0 &    35.0 &    20.0 &    57.0 &    37.0 &   28.0 &          32.95 \\
         & XLT &  66.67 &   24.0 &    39.0 &    50.0 &    53.5 &    57.0 &   33.5 &          46.24 \\
\hline
LLaMA-2-70B & Direct &  48.67 &   33.0 &    33.0 &    30.0 &    50.0 &    58.0 &   20.5 &          39.02 \\
         & ZH-CoT &  46.00 &   20.0 &    35.0 &    32.0 &    51.5 &    38.0 &   35.5 &          36.86 \\
         & EN-CoT &  46.00 &   82.0 &    53.0 &    34.0 &    51.5 &    64.0 &   48.5 &          54.14 \\
         & Translate-EN &  84.67 &   76.0 &    58.0 &    52.0 &    71.0 &    64.0 &   57.5 &          66.17 \\
         & XLT &  72.67 &   84.0 &    73.0 &    42.0 &    64.0 &    61.0 &   41.5 &          62.60 \\
\hline
Vicuna-7B-v1.5 & Direct &  15.33 &   22.0 &    33.0 &    12.0 &    49.5 &    22.0 &   16.5 &          24.33 \\
         & ZH-CoT &  52.00 &   50.0 &    50.0 &    10.0 &    50.5 &    53.0 &   31.0 &          42.36 \\
         & EN-CoT &  49.33 &   45.0 &    44.0 &    16.0 &    51.0 &    23.0 &   31.5 &          37.12 \\
         & Translate-EN &  76.00 &   58.0 &    47.0 &    36.0 &    66.5 &    57.0 &   37.5 &          54.00 \\
         & XLT &  45.33 &   64.0 &    37.0 &    26.0 &    58.5 &    52.0 &   32.5 &          45.05 \\
\hline
Vicuna-13B-v1.5 & Direct &  55.33 &   58.0 &    39.0 &    28.0 &    48.5 &    59.0 &   30.0 &          45.40 \\
         & ZH-CoT &  71.33 &   71.0 &    49.0 &    14.0 &    53.0 &    62.0 &   33.0 &          50.48 \\
         & EN-CoT &  66.00 &   82.0 &    42.0 &    38.0 &    65.0 &    55.0 &   40.5 &          55.50 \\
         & Translate-EN &  84.00 &   66.0 &    60.0 &    66.0 &    71.0 &    60.0 &   42.0 &          64.14 \\
         & XLT &  72.67 &   74.0 &    41.0 &    50.0 &    68.0 &    61.0 &   36.0 &          57.52 \\
\hline
ChatGLM3-6B & Direct &  44.00 &   33.0 &    57.0 &    42.0 &    63.0 &    80.0 &   38.0 &          51.00 \\
         & ZH-CoT &  34.00 &   69.0 &    71.0 &    28.0 &    75.5 &    63.0 &   34.0 &          53.50 \\
         & EN-CoT &  41.33 &   65.0 &    63.0 &    24.0 &    60.5 &    70.0 &   34.0 &          51.12 \\
         & Translate-EN &  66.67 &   59.0 &    70.0 &    42.0 &    66.5 &    71.0 &   39.5 &          59.24 \\
         & XLT &  52.67 &   58.0 &    70.0 &    46.0 &    66.0 &    66.0 &   42.5 &          57.31 \\
\hline
Baichuan2-7B & Direct &  46.00 &   20.0 &    47.0 &     8.0 &    58.0 &    35.0 &   38.0 &          36.00 \\
         & ZH-CoT &  55.33 &   65.0 &    54.0 &    26.0 &    60.5 &    59.0 &   29.0 &          49.83 \\
         & EN-CoT &  44.00 &   64.0 &    49.0 &    20.0 &    58.5 &    56.0 &   31.5 &          46.14 \\
         & Translate-EN &  73.33 &   59.0 &    48.0 &    28.0 &    64.0 &    54.0 &   36.5 &          51.83 \\
         & XLT &  48.67 &   18.0 &    49.0 &    34.0 &    56.0 &    50.0 &   23.0 &          39.81 \\
\hline
Baichuan2-13B & Direct &  64.00 &   17.0 &    55.0 &    20.0 &    58.0 &    37.0 &   23.5 &          39.21 \\
         & ZH-CoT &  77.33 &   74.0 &    58.0 &    40.0 &    71.0 &    61.0 &   39.0 &          60.05 \\
         & EN-CoT &  78.67 &   70.0 &    55.0 &    30.0 &    57.0 &    66.0 &   37.5 &          56.31 \\
         & Translate-EN &  73.33 &   68.0 &    51.0 &    36.0 &    61.5 &    61.0 &   42.0 &          56.12 \\
         & XLT &  70.67 &   75.0 &    49.0 &    42.0 &    69.5 &    61.0 &   31.0 &          56.88 \\
\hline
InternLM2-7B & Direct &  46.67 &   61.0 &    65.0 &    46.0 &    67.0 &    79.0 &   53.5 &          59.74 \\
         & ZH-CoT &  74.67 &   80.0 &    62.0 &    20.0 &    78.0 &    76.0 &   23.5 &          59.17 \\
         & EN-CoT &  72.00 &   87.0 &    70.0 &    44.0 &    76.0 &    73.0 &   38.5 &          65.79 \\
         & Translate-EN &  70.67 &   81.0 &    75.0 &    60.0 &    78.0 &    73.0 &   48.5 &          69.45 \\
         & XLT &  66.00 &   87.0 &    72.0 &    52.0 &    76.5 &    66.0 &   43.5 &          66.14 \\
\hline
InternLM2-20B & Direct &  81.33 &   54.0 &    78.0 &    50.0 &    63.5 &    46.0 &   48.0 &          60.12 \\
         & ZH-CoT &  82.67 &   83.0 &    61.0 &    14.0 &    74.5 &    72.0 &   27.0 &          59.17 \\
         & EN-CoT &  73.33 &   83.0 &    63.0 &    14.0 &    75.0 &    73.0 &   26.5 &          58.26 \\
         & Translate-EN &  82.00 &   84.0 &    89.0 &    40.0 &    76.0 &    68.0 &   46.5 &          69.36 \\
         & XLT &  87.33 &   84.0 &    80.0 &    70.0 &    79.0 &    70.0 &   47.0 &          73.90 \\
\hline
Yi-6B & Direct &  47.33 &   17.0 &    47.0 &    14.0 &    25.0 &    11.0 &   23.0 &          26.33 \\
         & ZH-CoT &  79.33 &   63.0 &    43.0 &    14.0 &    70.5 &    57.0 &   33.5 &          51.48 \\
         & EN-CoT &  68.67 &   56.0 &    53.0 &    32.0 &    55.5 &    27.0 &   41.0 &          47.60 \\
         & Translate-EN &  72.67 &   57.0 &    62.0 &    32.0 &    69.0 &    37.0 &   35.5 &          52.17 \\
         & XLT &  54.67 &   44.0 &    60.0 &    62.0 &    70.5 &    59.0 &   41.5 &          55.95 \\
\hline
Yi-34B & Direct &  82.67 &   67.0 &    85.0 &    58.0 &    53.5 &    45.0 &   64.0 &          65.02 \\
         & ZH-CoT &  88.67 &   92.0 &    87.0 &    56.0 &    89.0 &    70.0 &   47.5 &          75.74 \\
         & EN-CoT &  89.33 &   91.0 &    88.0 &    44.0 &    80.0 &    66.0 &   48.0 &          72.33 \\
         & Translate-EN &  78.00 &   85.0 &    83.0 &    48.0 &    76.5 &    64.0 &   54.5 &          69.86 \\
         & XLT &  88.00 &   88.0 &    86.0 &    70.0 &    93.5 &    60.0 &   58.0 &          77.64 \\
\hline
DeepSeek-7B & Direct &  47.33 &   24.0 &    35.0 &    14.0 &    22.0 &    41.0 &   17.5 &          28.69 \\
         & ZH-CoT &  68.00 &   76.0 &    47.0 &    50.0 &    72.5 &    59.0 &   32.5 &          57.86 \\
         & EN-CoT &  75.33 &   74.0 &    40.0 &    16.0 &    53.5 &    47.0 &   35.5 &          48.76 \\
         & Translate-EN &  72.67 &   59.0 &    45.0 &    32.0 &    60.0 &    57.0 &   38.0 &          51.95 \\
         & XLT &  58.00 &   28.0 &    38.0 &    16.0 &    51.5 &    35.0 &   29.5 &          36.57 \\
\hline
DeepSeek-67B & Direct &  37.33 &   83.0 &    18.0 &     2.0 &    39.5 &    49.0 &   37.0 &          37.98 \\
         & ZH-CoT &  90.00 &   95.0 &    86.0 &    22.0 &    88.0 &    73.0 &   39.0 &          70.43 \\
         & EN-CoT &  61.33 &   96.0 &    76.0 &    30.0 &    90.5 &    71.0 &   35.0 &          65.69 \\
         & Translate-EN &  90.67 &   87.0 &    81.0 &    52.0 &    81.0 &    54.0 &   58.5 &          72.02 \\
         & XLT &  86.00 &   93.0 &    72.0 &    60.0 &    93.0 &    64.0 &   46.0 &          73.43 \\
\hline
Qwen-7B & Direct &  52.67 &   38.0 &    54.0 &    38.0 &    56.5 &    67.0 &   40.0 &          49.45 \\
         & ZH-CoT &  74.67 &   78.0 &    69.0 &    50.0 &    72.5 &    55.0 &   36.0 &          62.17 \\
         & EN-CoT &  74.00 &   81.0 &    65.0 &    36.0 &    73.5 &    66.0 &   35.5 &          61.57 \\
         & Translate-EN &  73.33 &   71.0 &    65.0 &    46.0 &    70.5 &    66.0 &   41.0 &          61.83 \\
         & XLT &  74.67 &   64.0 &    69.0 &    48.0 &    67.0 &    46.0 &   32.0 &          57.24 \\
\hline
Qwen-14B & Direct &  70.00 &   58.0 &    82.0 &    36.0 &    78.0 &    55.0 &   47.5 &          60.93 \\
         & ZH-CoT &  84.00 &   83.0 &    83.0 &    44.0 &    84.5 &    71.0 &   40.0 &          69.93 \\
         & EN-CoT &  86.67 &   82.0 &    81.0 &    44.0 &    79.5 &    66.0 &   42.5 &          68.81 \\
         & Translate-EN &  86.67 &   72.0 &    85.0 &    48.0 &    78.0 &    64.0 &   48.5 &          68.88 \\
         & XLT &  80.00 &   79.0 &    83.0 &    48.0 &    79.0 &    65.0 &   45.0 &          68.43 \\
\hline
Qwen-72B & Direct &  88.00 &   63.0 &    85.0 &    56.0 &    83.5 &    78.0 &   65.5 &          74.14 \\
         & ZH-CoT &  94.00 &   92.0 &    93.0 &    64.0 &    93.0 &    71.0 &   63.5 &          81.50 \\
         & EN-CoT &  90.00 &   92.0 &    86.0 &    60.0 &    92.5 &    66.0 &   58.0 &          77.79 \\
         & Translate-EN &  91.33 &   87.0 &    89.0 &    54.0 &    81.5 &    63.0 &   64.0 &          75.69 \\
         & XLT &  92.67 &   70.0 &    91.0 &    66.0 &    91.5 &    66.0 &   50.5 &          75.38 \\
\bottomrule
\end{tabular}
\normalsize
    \caption{Accuracy of reasoning tasks in the \textbf{global} commonsense domain of CHARM.}
    \label{tab:gl_tasks_reasoning_res_19LLMsx5p}
\end{table*}

\begin{table*}
    \centering
\small
\begin{tabular}{llcccccccc}
\toprule
LLM & Prompt &  \textit{H} &  \textit{CA} &  \textit{LC} &  \textit{E}&  \textit{F}&  \textit{G} &  \textit{L} &   Avg. \\

\midrule
Random Baseline & - & 43.88 & 25.68 & 25.74 & 27.23 & 49.75 & 28.32 & 27.72 & 32.61 \\
\midrule

GPT-3.5-1106 & XLT &    80.85 &                        41.78 &                   63.24 &                                       49.30 &           80.88 &      52.21 &             55.45 &  60.53 \\
GPT-4-1106 & XLT &    92.55 &                        58.22 &                   86.76 &                                       73.24 &           86.27 &      71.68 &             67.27 &  76.57 \\
LLaMA-2-7B & XLT &    45.74 &                        16.44 &                   22.06 &                                       26.76 &           49.02 &      26.55 &             21.82 &  29.77 \\
LLaMA-2-13B & XLT &    51.06 &                        38.36 &                   39.71 &                                       36.62 &           57.84 &      46.02 &             30.91 &  42.93 \\
LLaMA-2-70B & XLT &    55.85 &                        38.36 &                   51.47 &                                       39.44 &           55.39 &      43.36 &             49.09 &  47.57 \\
Vicuna-7B-v1.5 & XLT &    49.47 &                        32.19 &                   33.82 &                                       38.03 &           50.49 &      44.25 &             32.73 &  40.14 \\
Vicuna-13B-v1.5 & XLT &    56.91 &                        38.36 &                   25.00 &                                       36.62 &           52.45 &      50.44 &             36.36 &  42.31 \\
\midrule
ChatGLM3-6B & ZH-CoT &    64.89 &                        37.67 &                   54.41 &                                       49.30 &           75.98 &      53.98 &             48.18 &  54.92 \\
Baichuan2-7B & ZH-CoT &    69.15 &                        45.21 &                   51.47 &                                       40.85 &           72.55 &      51.33 &             47.27 &  53.98 \\
Baichuan2-13B & ZH-CoT &    79.26 &                        43.84 &                   54.41 &                                       47.89 &           72.55 &      59.29 &             49.09 &  58.05 \\
InternLM2-7B & ZH-CoT &    76.60 &                        33.56 &                   61.76 &                                       49.30 &           75.49 &      49.56 &             45.45 &  55.96 \\
InternLM2-20B & ZH-CoT &    78.72 &                        40.41 &                   47.06 &                                       54.93 &           74.02 &      48.67 &             49.09 &  56.13 \\
Yi-6B & ZH-CoT &    61.70 &                        42.47 &                   54.41 &                                       38.03 &           75.00 &      44.25 &             36.36 &  50.32 \\
Yi-34B & ZH-CoT &    89.36 &                        56.16 &                   82.35 &                                       73.24 &           88.73 &      63.72 &             60.91 &  73.50 \\
DeepSeek-7B & ZH-CoT &    70.21 &                        37.67 &                   42.65 &                                       56.34 &           79.41 &      38.05 &             44.55 &  52.70 \\
DeepSeek-67B & ZH-CoT &    87.23 &                        55.48 &                   75.00 &                                       87.32 &           86.76 &      52.21 &             52.73 &  70.96 \\
Qwen-7B & ZH-CoT &    65.43 &                        42.47 &                   51.47 &                                       52.11 &           70.59 &      53.98 &             53.64 &  55.67 \\
Qwen-14B & ZH-CoT &    81.91 &                        62.33 &                   70.59 &                                       60.56 &           82.84 &      58.41 &             56.36 &  67.57 \\
Qwen-72B & ZH-CoT &    92.55 &                        61.64 &                   89.71 &                                       83.10 &           86.76 &      75.22 &             77.27 &  80.89 \\
\bottomrule
\end{tabular}
\normalsize
    \caption{Accuracy of reasoning questions on the 7 Chinese commonsense aspects of CHARM.}
\label{tab:cn_tasks_reasoning_7aspects_res_19LLMs}
\end{table*}

\begin{table*}
    \centering
\small
\begin{tabular}{lrrrr}
\toprule
LLM &  \# Original &  Original Acc. &  \# Retained &  Retained Acc. \\
\midrule
DeepSeek-67B        &         500 &           84.6 &         409 &          87.04 \\
Qwen-72B            &         500 &           84.2 &         402 &          84.33 \\
GPT-4-1106          &         500 &           82.8 &         350 &          86.86 \\
Yi-34B              &         500 &           82.8 &         355 &          89.86 \\
Qwen-14B            &         500 &           76.0 &         331 &          80.97 \\
InternLM2-20B       &         500 &           71.6 &         329 &          78.42 \\
GPT-3.5-1106        &         500 &           69.8 &         226 &          81.86 \\
InternLM2-7B        &         500 &           68.2 &         282 &          78.37 \\
DeepSeek-7B         &         500 &           68.0 &         338 &          75.74 \\
Baichuan2-13B       &         500 &           67.2 &         320 &          76.56 \\
Baichuan2-7B        &         500 &           63.6 &         337 &          64.69 \\
ChatGLM3-6B     &         500 &           62.4 &         258 &          65.12 \\
Qwen-7B             &         500 &           62.0 &         323 &          63.78 \\
Yi-6B               &         500 &           60.8 &         285 &          68.07 \\
LLaMA-2-70B         &         500 &           49.8 &         107 &          53.27 \\
LLaMA-2-13B         &         500 &           49.6 &          70 &          55.71 \\
Vicuna-13B-v1.5 &         500 &           47.6 &         106 &          49.06 \\
Vicuna-7B-v1.5  &         500 &           45.0 &          62 &          40.32 \\
LLaMA-2-7B          &         500 &           43.8 &          49 &          46.94 \\
\bottomrule
\end{tabular}
\normalsize
    \caption{Filtering Reasoning questions based on Mono-LLM-Memorization (FRMM) on the \textit{MRI} tasks.}
\label{tab:filter-rea-based-on-mem-19LLMs}
\end{table*}

\begin{table*}
    \centering
\small

\begin{tabular}{lr}
\toprule
LLM &   Final Score \\
\midrule
GPT-4-1106          &  21.60 \\
Yi-34B              &  19.52 \\
Qwen-72B            &  18.43 \\
DeepSeek-67B        &  18.25 \\
GPT-3.5-1106        &  12.65 \\
Qwen-14B            &  10.80 \\
InternLM2-20B       &  10.01 \\
InternLM2-7B        &   7.30 \\
Baichuan2-13B       &   5.71 \\
DeepSeek-7B         &   5.17 \\
Baichuan2-7B        &  -0.77 \\
ChatGLM3-6B    &  -3.11 \\
Qwen-7B             &  -4.04 \\
Yi-6B               &  -6.22 \\
LLaMA-2-13B         & -15.31 \\
LLaMA-2-70B         & -22.69 \\
Vicuna-13B-v1.5 & -22.92 \\
Vicuna-7B-v1.5  & -26.25 \\
LLaMA-2-7B          & -28.13 \\
\bottomrule
\end{tabular}

\normalsize
    \caption{Final results of the Memorization-Independent Battles among LLMs (MIB) on the \textit{MRI} tasks.}
\label{tab:final-res-MIB-19LLMs}
\end{table*}

\end{document}